\documentclass[journal,onecolumn]{IEEEtran}

\usepackage{epsfig}
\usepackage{graphicx}
\usepackage{amsmath}
\usepackage{amssymb}
\usepackage{commath}
\usepackage{bbding}
\usepackage{cite}
\usepackage{cleveref}
\usepackage{booktabs,multirow}
\usepackage{times}

\usepackage{mathrsfs}

\usepackage{mathtools}

\DeclareMathOperator*{\med}{Median}

\begin{document}

%--------------------------------------------------------------------------------------------- 
%--------------------------------------------------------------------------------------------- 
%--------------------------------------------------------------------------------------------- 
\title{\huge View-Invariant Recognition of Action Style Self-Dissimilarity}
%
%
%--------------------------------------------------------------------------------------------- 
%--------------------------------------------------------------------------------------------- 

\author{Yuping~Shen\thanks{Yuping Shen was with the Department of Computer Science, University of Central Florida, Orlando, FL, 32816 USA at the time this project was conducted. (e-mail: 
ypshen@cs.ucf.edu).} and Hassan~Foroosh\thanks{Hassan Foroosh is with the Department of Computer Science, University of Central Florida, Orlando, FL, 32816 USA (e-mail: foroosh@cs.ucf.edu).}
}

\maketitle

%--------------------------------------------------------------------------------------------- 
%--------------------------------------------------------------------------------------------- 
\begin{abstract}
Self-similarity was recently introduced as a measure of inter-class congruence for classification of actions. Herein, we investigate the dual problem of intra-class dissimilarity for classification of action styles. We introduce self-dissimilarity matrices that discriminate between same actions performed by different subjects regardless of viewing direction and camera parameters. We investigate two frameworks using these invariant style dissimilarity measures based on Principal Component Analysis (PCA) and Fisher Discriminant Analysis (FDA). Extensive experiments performed on IXMAS dataset indicate remarkably good discriminant characteristics for the proposed invariant measures for gender recognition from video data.
\end{abstract}


%--------------------------------------------------------------------------------------------- 
%--------------------------------------------------------------------------------------------- 
\section{Introduction}

Human action recognition from video data data has a wide range of applications in areas such as surveillance and image retrieval \cite{Junejo_Foroosh_2008,Sun_etal_2012,junejo2007trajectory,sun2011motion,Ashraf_etal2012,sun2014feature,Junejo_Foroosh2007-1,Junejo_Foroosh2007-2,Junejo_Foroosh2007-3,ashraf2012motion,ashraf2015motion,sun2014should}, image annotation \cite{Tariq_etal_2017,Tariq_etal_2017_2,tariq2013exploiting,tariq2015feature,tariq2014scene}, video post-production and editing \cite{balci2006real,moore2008learning,alnasser2006image,Alnasser_Foroosh_rend2006,fu2004expression,balci2006image},
and self-localization \cite{Junejo_etal_2010,Junejo_Foroosh_2010,Junejo_Foroosh_solar2008,Junejo_Foroosh_GPS2008,junejo2008gps}, to name a few.

The literature on human action recognition from video data includes both monocular and multiple view methods \cite{Shen_Foroosh_2009,Ashraf_etal_2014,Sun_etal_2015,shen2008view,sun2011action,ashraf2014view,shen2008action,shen2008view-2,ashraf2010view,boyraz122014action,Shen_Foroosh_FR2008,Shen_Foroosh_pose2008,ashraf2012human}. Often, multiple view methods are designed to tackle viewpoint invariant recognition \cite{Shen_Foroosh_2009,Ashraf_etal_2014,shen2008view,ashraf2014view,shen2008view-2,ashraf2010view,Shen_Foroosh_FR2008,ashraf2012human}, although such methods may require calibration across views \cite{Junejo_etal_2011,junejo2006dissecting,junejo2007robust,junejo2006robust,Junejo_Foroosh_calib2008,Junejo_Foroosh_PTZ2008,Junejo_Foroosh_SolCalib2008,Ashraf_Foroosh_2008,Junejo_Foroosh_Givens2008,Balci_Foroosh_metro2005},  image registration \cite{Foroosh_etal_2002,Foroosh_2005,Balci_Foroosh_2006,Balci_Foroosh_2006_2,Alnasser_Foroosh_2008,Atalay_Foroosh_2017,Atalay_Foroosh_2017-2,shekarforoush1996subpixel,foroosh2004sub,shekarforoush1995subpixel,balci2005inferring,balci2005estimating,Balci_Foroosh_phase2005,Foroosh_Balci_2004,foroosh2001closed,shekarforoush2000multifractal,balci2006subpixel,balci2006alignment,foroosh2004adaptive,foroosh2003adaptive}, or tracking across views \cite{Shu_etal_2016,Milikan_etal_2017,Millikan_etal2015,shekarforoush2000multi,millikan2015initialized}. There are also methods that rely on human-object interaction \cite{prest2012weakly,yao2011human,yao2012recognizing}, which often require identifying image contents other than humans \cite{liu2015sparse,wang2016factorized,Cakmakci_etal_2008,Cakmakci_etal_2008_2,Lotfian_Foroosh_2017,Morley_Foroosh2017,Ali-Foroosh2016,Ali-Foroosh2015,Einsele_Foroosh_2015,ali2016character,Cakmakci_etal2008,damkjer2014mesh}.
Other preprocessing steps that may be needed include image restoration \cite{Foroosh_2000,Foroosh_Chellappa_1999,Foroosh_etal_1996,berthod1994reconstruction,shekarforoush19953d,lorette1997super,shekarforoush1998multi,shekarforoush1996super,shekarforoush1995sub,shekarforoush1999conditioning,shekarforoush1998adaptive,berthod1994refining,shekarforoush1998denoising,bhutta2006blind,jain2008super,shekarforoush2000noise,shekarforoush1999super,shekarforoush1998blind},
or scene modeling \cite{Junejo_etal_2013,bhutta2011selective,junejo1dynamic,ashraf2007near}.

In this paper, we look at a very specific problem of determining stylistic differences in an action performed by different groups of people of people, e.g. stylistic difference due to age or gender differences. Human action style analysis is an important area in interpreting activities, which is motivated by the need for various applications, such as surveillance system, ergonomic evaluation, etc. At the very limit, of course, an individual could be considered as a category of their own, in which case this problem reduces to determining the identity of the individual, e.g. in gate recognition.\\

The problem of action style analysis is related to action recognition problem in that, action recognition systems aim at finding the features that distinguish different actions, while in action style analysis, stylistic features, e.g., stride parameters of walking gaits, are extracted from instances of the same action to reflect the style variations of individuals, or groups of individuals. 

\section{Related Work}
It has been proven that humans can recognize actions from limited types of input such as point lights and low quality video \cite{bobick2001rhm,johansson1973vpb}.
%As introduced in chapter \ref{ScChapter4}, there has been a great variety of work in computer vision on detecting, tracking and recognizing human actions.
Even with such limited information, we are capable to differentiate stylistic action differences, such as the gender \cite{barclay1978tas,pollick2002estimating} and age \cite{davis2001vcc} of a walking person. There has been several recent work on the study of action style variation in computer vision. Wilson and Bobick \cite{wilson1999phm} use a Parameterized-HMM to model spatial pointing gestures by adding a global variation paramters in the output probabilities of the HMM states. In \cite{tenenbaum9fss} Tenenbaum et al. use a bilinear model to separate perceptual content and style parameters. Davis \cite{davis2001vcc} proposed an approach to determine age of people based on variations in relative stride length and stride frequency over various walking speeds. In \cite{davis2002aar} Davis and Taylor use regularities in walking to classify typical from atypical gaits. Davis et al. \cite{davis2004etm} presented a three-mode (body pose, time, and style) expressive-feature model for representing and recognizing performance styles of human actions. The application of style analysis in computer animation for generating new animation styles of human motion have also been reported in \cite{unuma1995fpe,brand2000sm,vasilescu2002hms,vasilescu2002mai,davis2002efm,chi2000eme}.

Gait recognition is another problem related to action style analysis, which has very important implications for different domains such as surveillance, medical diagnosis, etc. It is based on the widely held belief that humans can distinguish between gait patterns of different individuals, by examining gait properties such as stride length, bounce, rhythm, and speed, etc. An early report on the ability to recognize people from gait was presented by Beardsworth and Buckner \cite{beardsworth1981aro}. They showed that the ability to recognize oneself from point-light features is greater than that of recognizing others, a surprising result. The ability of people to identify others using gait information alone has also been supported by the studies of Stevenage et al. \cite{stevenage1999vag} and Schollhorn et al. \cite{schollhorn2002iiw}.
Existing approaches in gait recognition can be broadly grouped into two categories: model based approaches \cite{lee2002gar,troje2002dbm,murray1964wpn} and model free approaches  \cite{benabdelkader2002mbr,huang1999rhg,little1998rpt,kale2003gah,sundaresan2003hmm,rabiner1989thm,tolliver2003gse}.
The model based methods assume a {\em priori} parameterized model and use these parameters as identify features of human. They try to fit the model to the 2D image sequences, and when the model and images are matched, the feature correspondence is automatically achieved. One example of model based approach is proposed in \cite{lee2002gar}, where a model consists of severa ellipses is used, and parameters of these ellipses such as their centroid and eccentricity are used as human identity features.
Troje \cite{troje2002dbm} proposed to determine the gender of walkers from trajectories of projection coefficients of body pose.
Murray \cite{murray1964wpn} model the hip rotation angle as a simple pendulum, and approximate its motion by simple harmonic motion. A similar work is also reported by Cunado et al. \cite{bobick2001gea}, who use an articulated pendulum-like motion model and extract a gait signiture by fitting the motion of the thighs to it.
The model free approaches can be further divided as deterministic or stochastic methods. Examples of deterministric methods include the work of Benabdelkader et al. \cite{benabdelkader2002mbr}, where the image self-similarity plot is used as a gait feature. Huang et al. \cite{huang1999rhg} use optical flow to derive a motion image sequence for a walk cycle, and apply principle components analysis to silhouettes to derive so-called eignen gaits. Little and Boyd \cite{little1998rpt} extract frequencey and phase features from moments of of the motion image, and use template matching to identify people by their gait. Kale et al. \cite{kale2003gah} extract gait features from width vectors, velocity profile etc. and use the sequence of feature vectors to represent gait. They then use dynamic time-warping approach to match two gait sequence to identify people. Other deterministic approaches also include \cite{kale2003gah,phillips2002brc,collins2002sbh}. Examples of stochastic methods include the work of Sundaresan \cite{sundaresan2003hmm}, where an HMM \cite{rabiner1989thm} is used to represent the gait of each individual. Another approach based on HMM is also reported in \cite{lee2003lpm}. Tolliver et al. \cite{tolliver2003gse} extract shape from a cluster of similar pose obtained from a spectral partitioning framework, and use it to identify different individuals.

Like in other problems of human motion analysis, view invariance is an important requirement in gait recognition. However, only a few papers in the literature take into account view invariance . Shakhnarovich et al. \cite{shakhnarovich2001ifa} propose an approach that integrates face and gait recognition from multiple views. As in other works that use multiple view data, their approach is limited to the number of views being used and is not ``truly'' view-invariant. Kale et al.\cite{kale2003tvi} propose a view-invariant method for the case when the person is far from the camera. They synthesize a side view from any other arbitrary view using a single camera, and apply methods based on side view of walking to solve the gait recognition problem.

\section{Action Style Analysis Using Homographies}

In a recent work, Shen et al. \cite{Shen2008,Shen2009} suggested that the non-rigid motion of an articulated body can be decomposed into rigid motions of planes given by triplets of points corresponding to the joints of the articulated body. This essentially implies that an articulated non-rigid motion can be described by a set of homographies. As a result the non-linear problem of modeling the motion of an articulated body can be reduced to a set of linear problems in terms of motions of a collection of planes associated with point triplets. Their work focused on inter-class classification using invariants associated with these homographies. Herein, we consider a similar decomposition, but will focus on the dual problem of intra-class classification. For example, in the case of two walking sequences by a male and a female subject, the motion of the body point triplet that includes the hip, knee and the foot usually appears different in the two sequences, due to different styles of swagger and body swing in male and female subjects. Therefore, it should be possible to derive invariant style features from the motion of such body point triplets for analysis of intra-class differences, e.g. for applications such as gender identification from video data.

%The key idea here is that while inter-class variations are determined by the entire set of homographies associated with body joints, the intra-class variations are reflected only in a subset of these homographies. For example, in the case of two walking sequences by a male and a female subject, the motion of the body point triplet that includes the hip, knee and the foot usually appears different in the two sequences, due to different styles of swagger and body swing in male and female subjects. Therefore, it should be possible to derive invariant style features from the motion of such body point triplets for analysis of intra-class differences, e.g. for applications such as gender identification from video data.

The first step is alignment of the target sequences to a reference sequence, which is discussed in the next section.

\begin{figure}[htb]
\centering
\includegraphics[width=30mm]{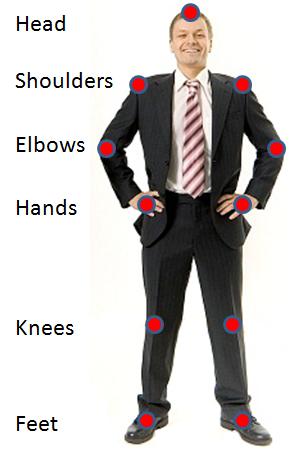}
\caption{Body points used for representing human body.}
\label{fig:body}
\end{figure}

\subsection{Action Sequence Alignment}
We call a camera view of a person's body motion from one body pose to another, a pose transition. Our action sequence alignment is based on aligning pose transitions of two actions viewed by two different cameras. We start by a view-invariant similarity measure proposed in \cite{Shen2008,Shen2009}: Suppose two different subjects perform actions that are viewed by two different cameras. Any triplet of body points in one view and the corresponding triplet of body points in the second view define a homography ${\bf H}_1$ between the two cameras. After transition of the triplets to a new position in space a second homography ${\bf H}_2$ would be induced. If the triplet motions by the two subjects differ only up to a similarity transformation, then the two homographies would be consistent with the fundamental matrix, and as a result the cross-homography defined by ${\bf H} = {\bf H}_1{\bf H}_2^{-1}$ would reduce to a homology. In other words two of the eigenvalues of ${\bf H}$ would be equal, e.g. $\sigma_1=\sigma_2$. The equality of the two eigenvalues of the cross-homography ${\bf H}$, would thus provide a similarity measure for the motion of the two body point triplets. Furthermore, this measure is invariant to viewing directions and the camera parameters. Given $11$ body points as shown in Figure 1, there are $165$ such triplets, all of which could be used to provide a combined measure of similarity of two actions.

Using this invariant measure, we define the following median absolute deviation estimator (MAD) as a measure of similarity of two pose transitions:
\begin{eqnarray}
\mathscr{M}(\mathcal{P},\mathcal{Q}) &=& \med_{i=1,..,N}\left|\frac{\sigma_{i1}}{\sigma_{i2}}-\mu\right| \label{eq:costf2}
\end{eqnarray}
where $\mu$ is the expected value of the ratio of the two closest eigenvalues over a subset of $N$ triplets. Since in our problem, we are only interested in intra-class alignment, $\mu$ can be set to one. \\

Suppose now we are given a target sequence of $m$ pose transitions $\mathcal{T}_j$, $j=1,...,m$ and a reference sequence of $n$ pose transitions $\mathcal{R}_{j'}$, $j'=1,...,n$.  The optimal pose transition $\mathcal{T}_j^*$, in the target sequence that best matches a given pose transition $\mathcal{R}_{j'}$ in the reference sequence can be obtained by:
\begin{equation}
\mathcal{T}_j^*= \arg\min_j \mathscr{M}(\mathcal{T}_j,\mathcal{R}_{j'})
\end{equation}
In order to find the optimal alignment $\psi: \mathcal{T}\rightarrow \mathcal{R}$ between the two sequences, we build the following matching error matrix
\begin{eqnarray}
{\bf P}=[p_{ij}],\;\;\;\;\;\;\mbox{where}\; p_{ij}=\mathscr{M}(\mathcal{T}_j,\mathcal{R}_{j'})
\end{eqnarray}
The problem is clearly well suited for dynamic programming, and hence the solution is found as the path in the error matrix that minimizes the cumulative error.
%Now, let $\psi: \mathcal{T}\rightarrow \mathcal{R}$ be a mapping that aligns a target sequence $\mathcal{T}$ of $m$ pose transitions with a reference sequence $\mathcal{R}$ of $n$ pose transitions.

Once a sequence is aligned to the reference sequence in a class of actions, the next question to answer is how dissimilar it is to the action class representative, i.e. the reference sequence. We will show in the next section that the dissimilarities are reflected in the motion patterns of planes defined by body point triplets, and can be measured in terms of the matching errors with respect to the reference sequence that represents our action class. Typically an intra-class classification is a much harder problem than an inter-class classification. Our approach has two important desirable characteristics:
\begin{enumerate}
\item Our error measures are based on the absolute deviation, and hence by design are meant to maximize discriminative power of our classifier.
\item Our classifier is invariant to camera parameters and orientations, and hence can rely on a much smaller set of training or reference sequences.
\end{enumerate}
In the next section, we introduce our intra-class dissimilarity measures and demonstrate their power.

\subsection{Self-Dissimilarity}\label{sec:styleFeature}
By self dissimilarity we mean how dissimilar is an instance of an action relative to its class representative. Once a target sequence is aligned to a reference sequence as described above, for every pose transition $\mathcal{T}_j$ in the target sequence we will have the corresponding pose transition $\mathcal{R}_{j'}$ in the reference sequence. We next build two matrices that reflect the differences in action style in the target sequence $\mathcal{T}$ compared with the reference sequence:
\begin{description}
\item[Triplet Deviation Matrix (TDM)]: We construct a matrix $\mathbf{M}_\psi$ as follows:
\begin{equation}
\mathbf{T} = \left[ t_{ij}\right], \mbox{where} t_{ij}=\left|\frac{\sigma_{i1}}{\sigma_{i2}}-\mu\right|_j
\end{equation}
where the subscript $j=1,...,m$ are pose transition indices and $i=1,...,N$ are the triplet indices.
%\begin{equation}
%\mathbf{T} = \left[ \begin{array}{cccc}
%{\bf t}_1 & {\bf t}_2 & \dots {\bf t}_m \\
%\end{array}\right]
%\end{equation}

Therefore each column of $\mathbf{T}$ is an $N$-vector containing the absolute deviation of all triplets for the matched pose transition $\mathcal{T}_j$. On the other hand, each row $i$ of the matrix corresponds to the absolute deviations of the triplet $i$ across all matching pose transitions. \\

%$\mathbf{T}_\psi$ reflects the variations of all triplets in $\mathcal{T}$ compared to the reference $\mathcal{R}$. A peak or valley in $\mathbf{T}_\psi$ means the motion of corresponding triplets at the specific poses appear quite different in $\mathcal{T}$ from those in $\mathcal{R}$. To reduce the effect of outliers introduced by degenerate or close-to-degenerate triplets, we apply a median filter to $\mathbf{T}_\psi$.

%We then normalize this matrix to an $N\times N$ matrix b$\hat{\mathbf{M}_e}$ to $n\times n$ matrix $\mathbf{M}_{tv}$, which is named as \textit{Triplet Variation Matrix},  by applying a x-direction linear interpolation on $\hat{\mathbf{M}_e}$.
%In theory, the elements o fa $EM$ built on two identical sequences are all equal to $\tau$

\item[Pose Deviation Matrix (PDM)]: For aligning the target sequence $\mathcal{T}$ to the reference sequence $\mathcal{R}$ we applied dynamic programming on the matrix ${\bf P}$. In this matrix each element is the dissimilarity error of pose $i$ of $\mathcal{R}$ and pose $j$ of $\mathcal{T}$. In a sense, the elements of ${\bf P}$ are a measure of correlation between pose transitions in the two sequences. When $\mathcal{T}$ and $\mathcal{R}$ are the same sequence, ${\bf P}$ is a special case of the self-similarity matrix, which was used for action recognition in \cite{junejo2008cross}. The patterns in ${\bf P}$ represent the characteristic features of the target sequence, and can be used to describe the style deviation of the target sequence from the reference sequence.
\end{description}
TDM and PDM describe the stylistic deviations of a sequence from the reference in two different ways: TDM captures localized low-level (body point triplet level) style deviations, while PDM captures a global motion style deviation by looking at the whole body pose.

In the following sections, we will discuss the properties of TDM and PDM. Without loss of generality, we study the action of kicking as an example. In our study, we use the IXMAS dataset \cite{weinland2006fva} in which videos of 13 actions are captured under 5 cameras, and each action is performed by 11 actors for 3 times/instances. For kicking action, we randomly chose the ``bao1'' sequence from camera 2 as the reference sequence.
\begin{figure*}[htb]
\centering
\includegraphics[width=170mm]{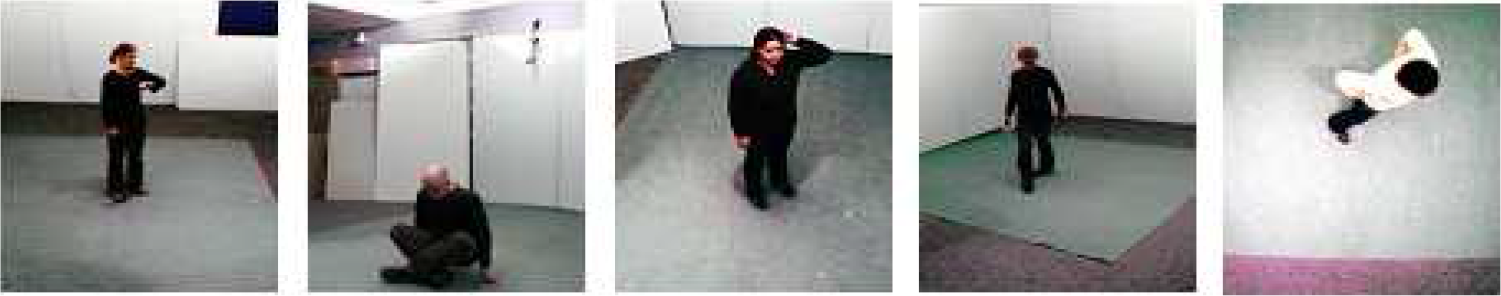}
\caption{Example sequences in IXMAS dataset.}
\label{fig:ixmas}
\end{figure*}

\subsubsection{Deviations of Individual Subjects}
\begin{figure}[h]
\centering
\begin{tabular}{ccc}
\includegraphics[width=55mm]{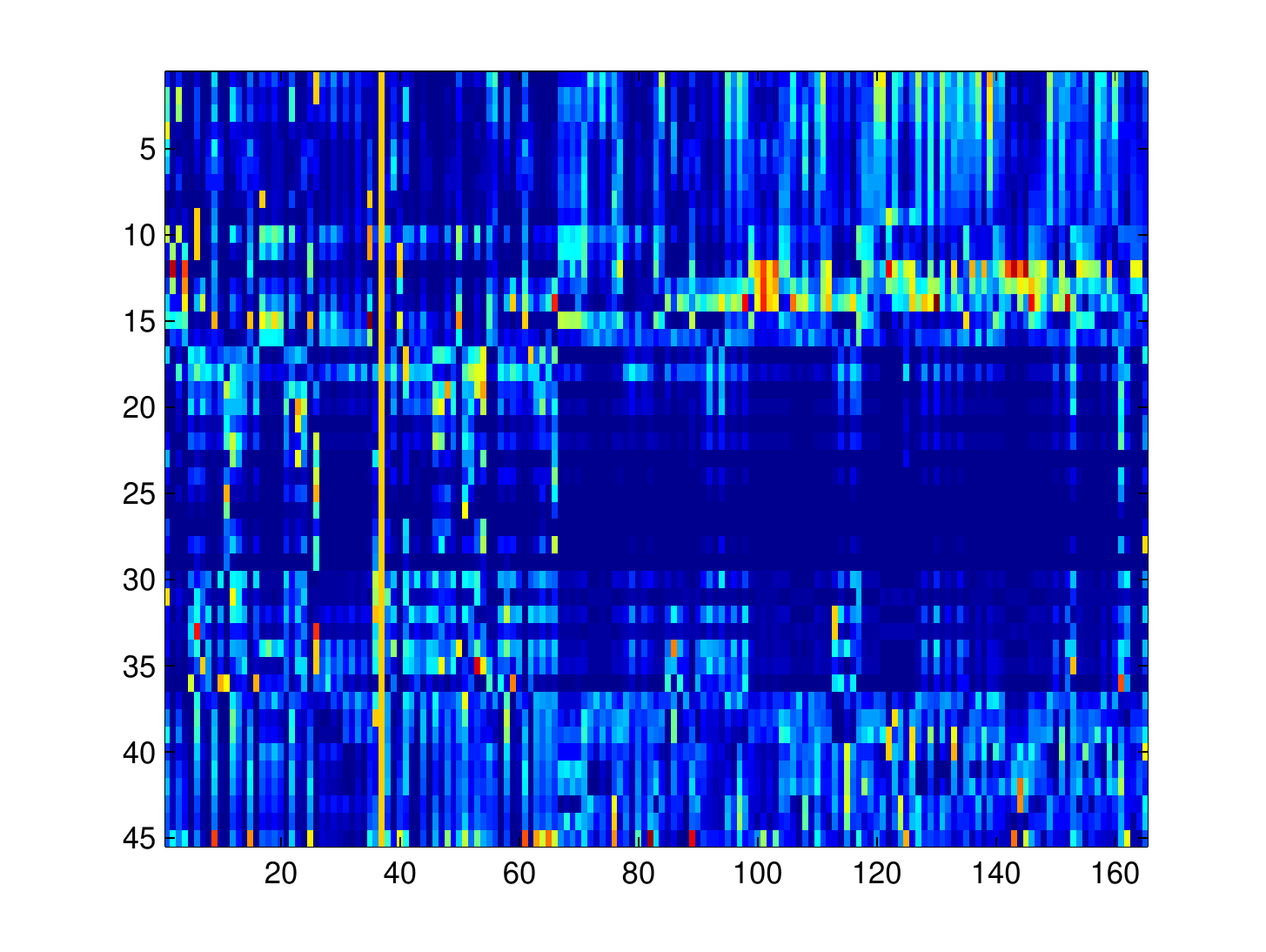} &
\includegraphics[width=55mm]{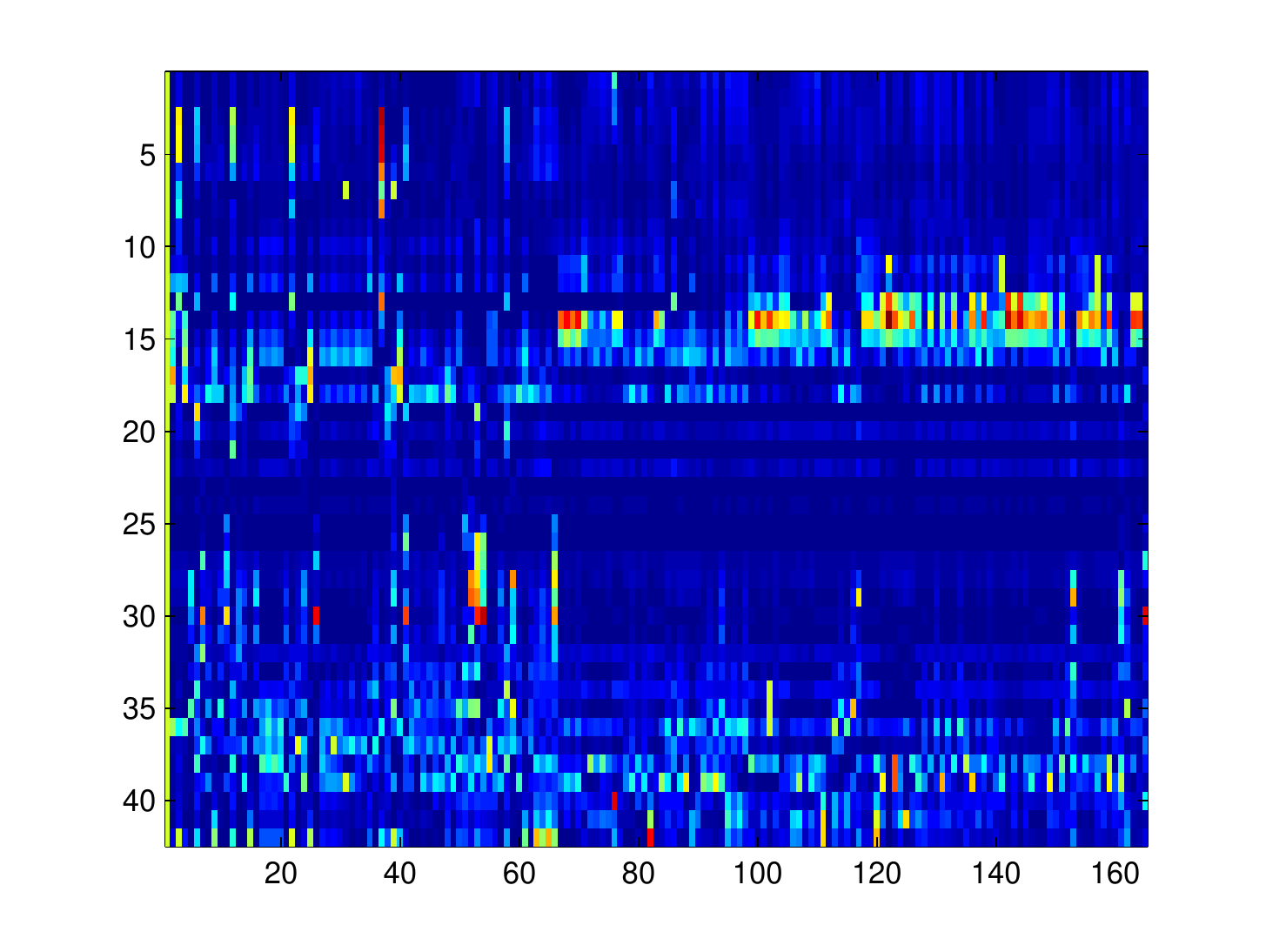} & \includegraphics[width=55mm]{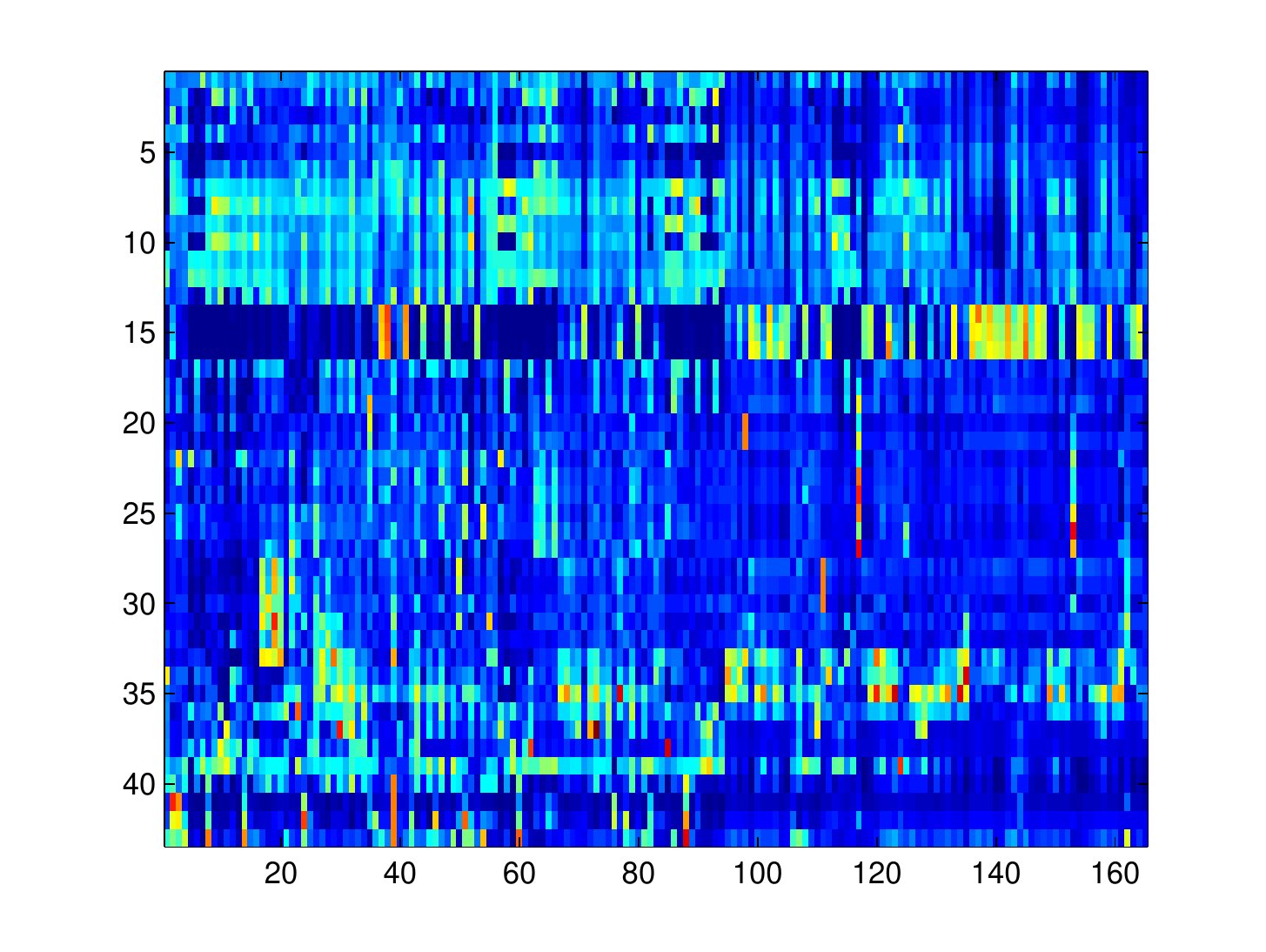} \\
(a)  & (b) & (c) \\
\end{tabular}
\caption{TDMs of sequences performed by different actors: (a) is associated to actor ``Florian'', (b) to ``Nicolas'', and (c) to ``Srikumar''}
\label{fig:TDM}
\end{figure}

\begin{figure}[htb]
\centering
\begin{tabular}{ccc}
\includegraphics[width=55mm]{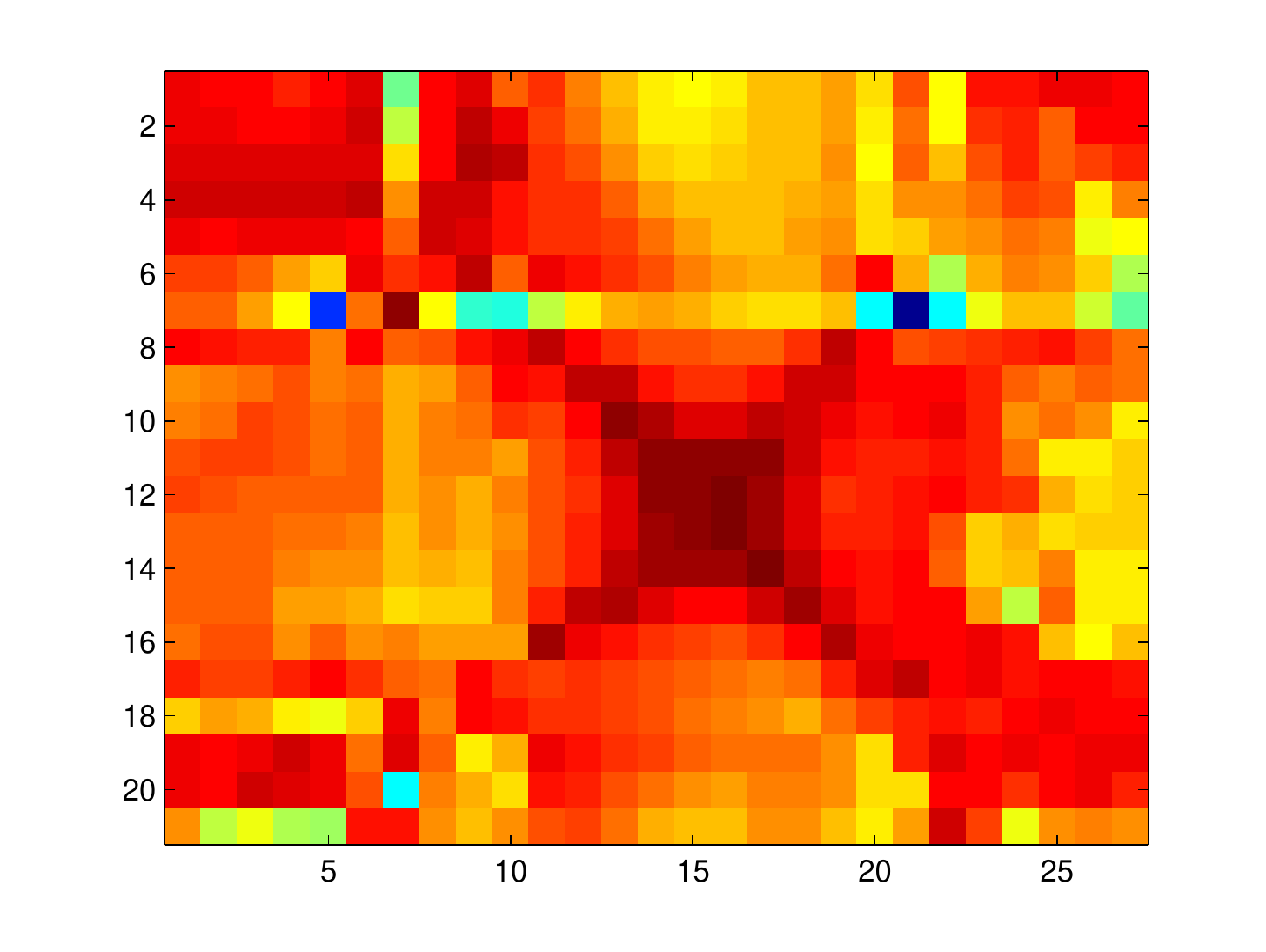} &
\includegraphics[width=55mm]{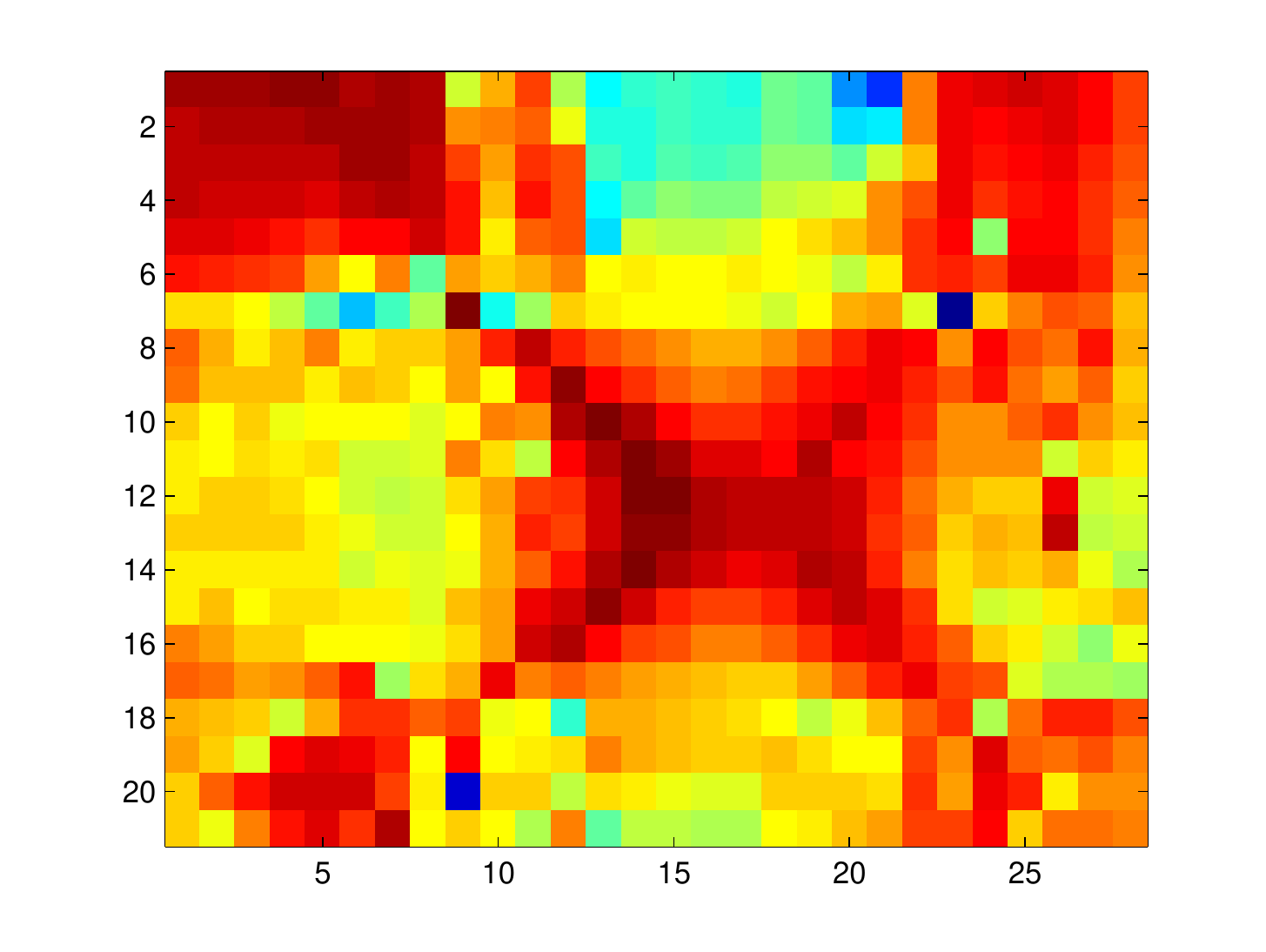} & \includegraphics[width=55mm]{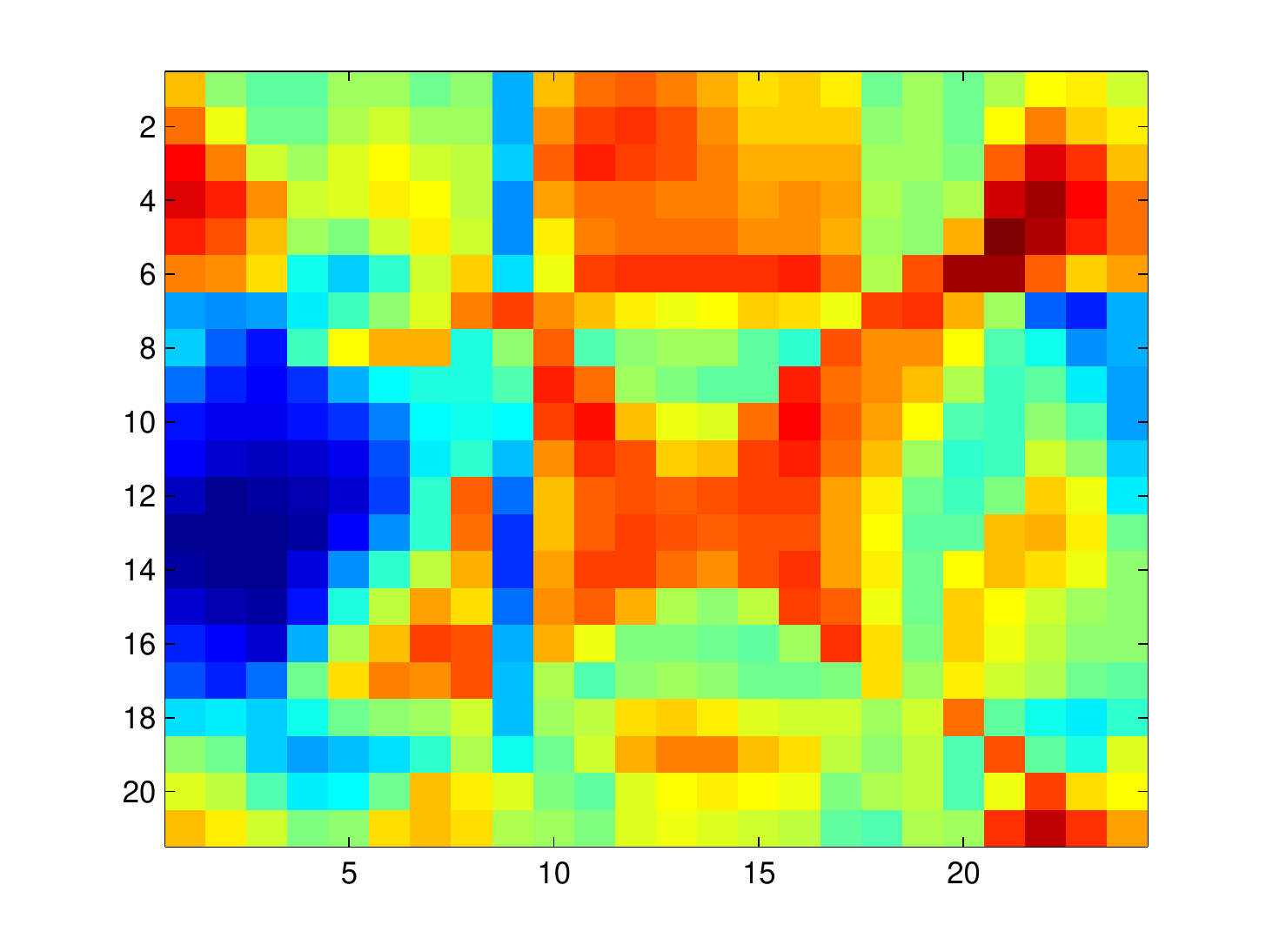} \\
(a)  & (b) & (c) \\
\end{tabular}
\caption{PDMs of sequences performed by different actors: (a) is associated to actor ``Florian'', (b) to ``Nicolas'', and (c) to ``Srikumar''}
\label{fig:PDM}
\end{figure}

We selected 3 kicking sequences performed by 3 actors in the data set, aligned them to the reference sequence and computed the corresponding TDM and PDM for each sequence (see Figures \ref{fig:TDM} and \ref{fig:PDM}).

As shown in Figures \ref{fig:TDM} and \ref{fig:PDM}, the TDM and PDM have different patterns in the three sequences. The different locations of peaks and valleys in the TDM plots suggest that the corresponding triplets move differently at various time slots in these sequences. The PDM also provides a good indication of different styles in these sequences, although their patterns in the diagonal look similar since they are intra-class measures.

\subsubsection{View Invariance}
\begin{figure*}[htb]
\centering
\begin{tabular}{cccc}
\hspace*{-1cm}\includegraphics[width=45mm]{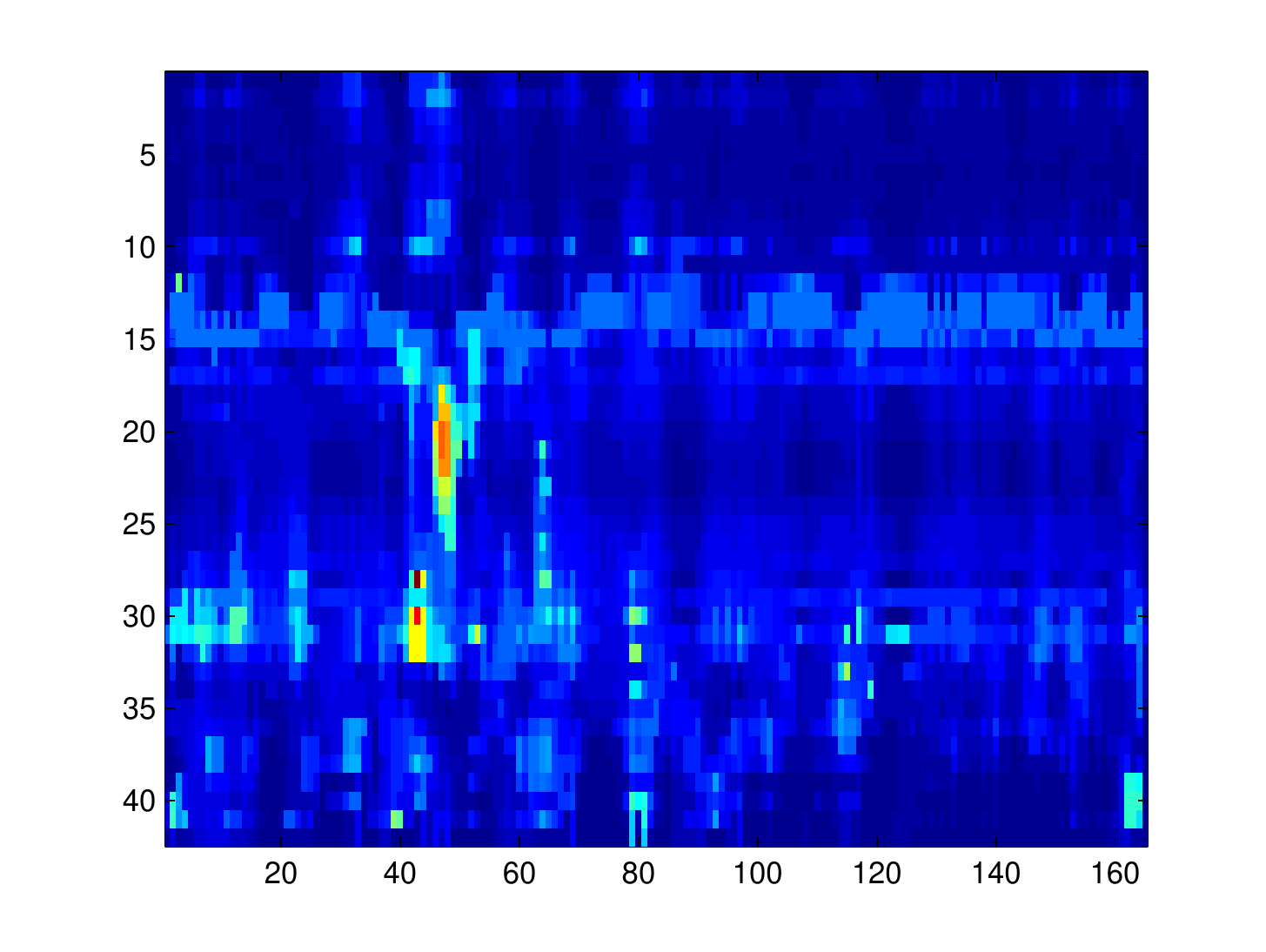} &%
\includegraphics[width=45mm]{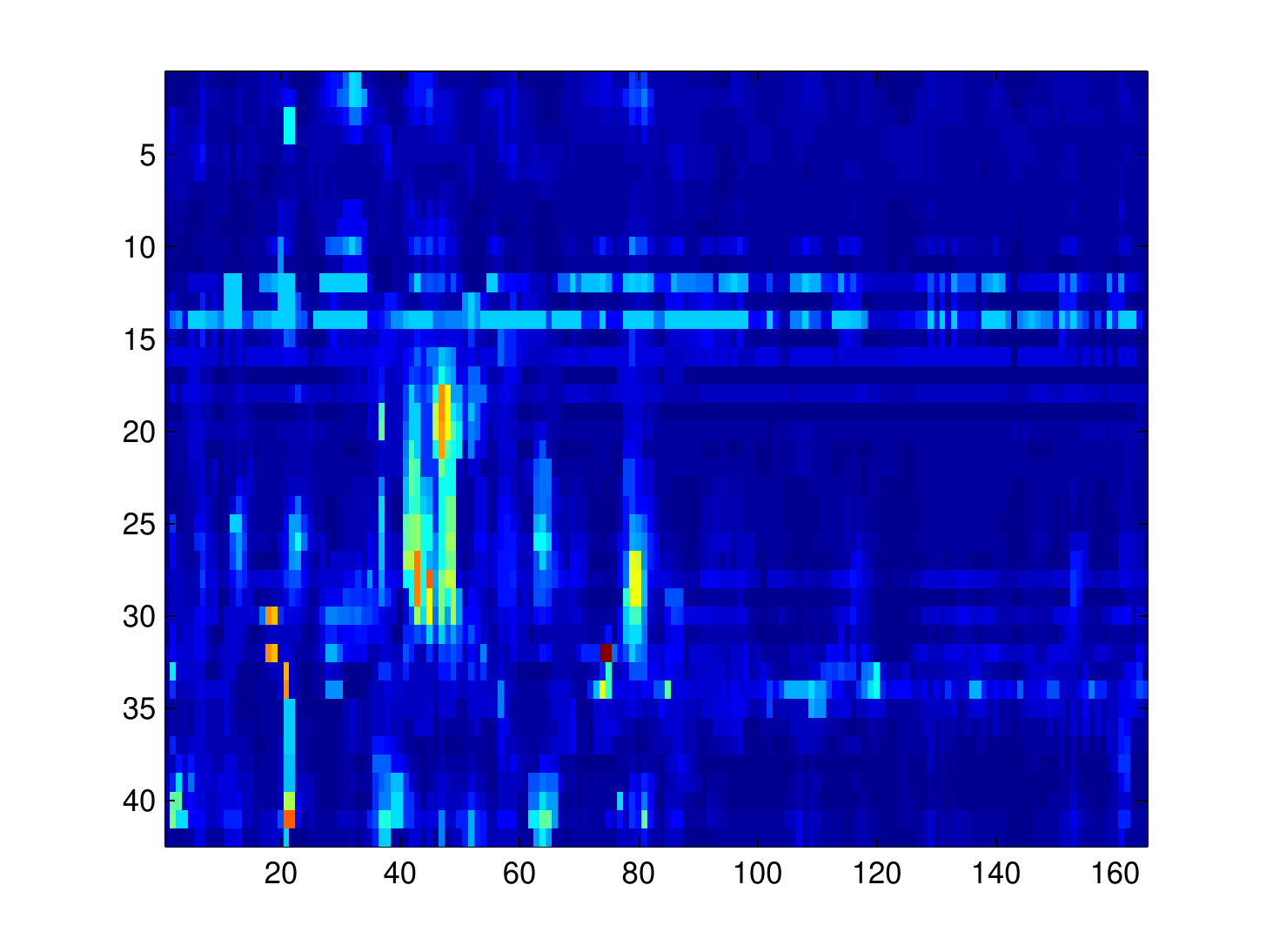} &%
\includegraphics[width=45mm]{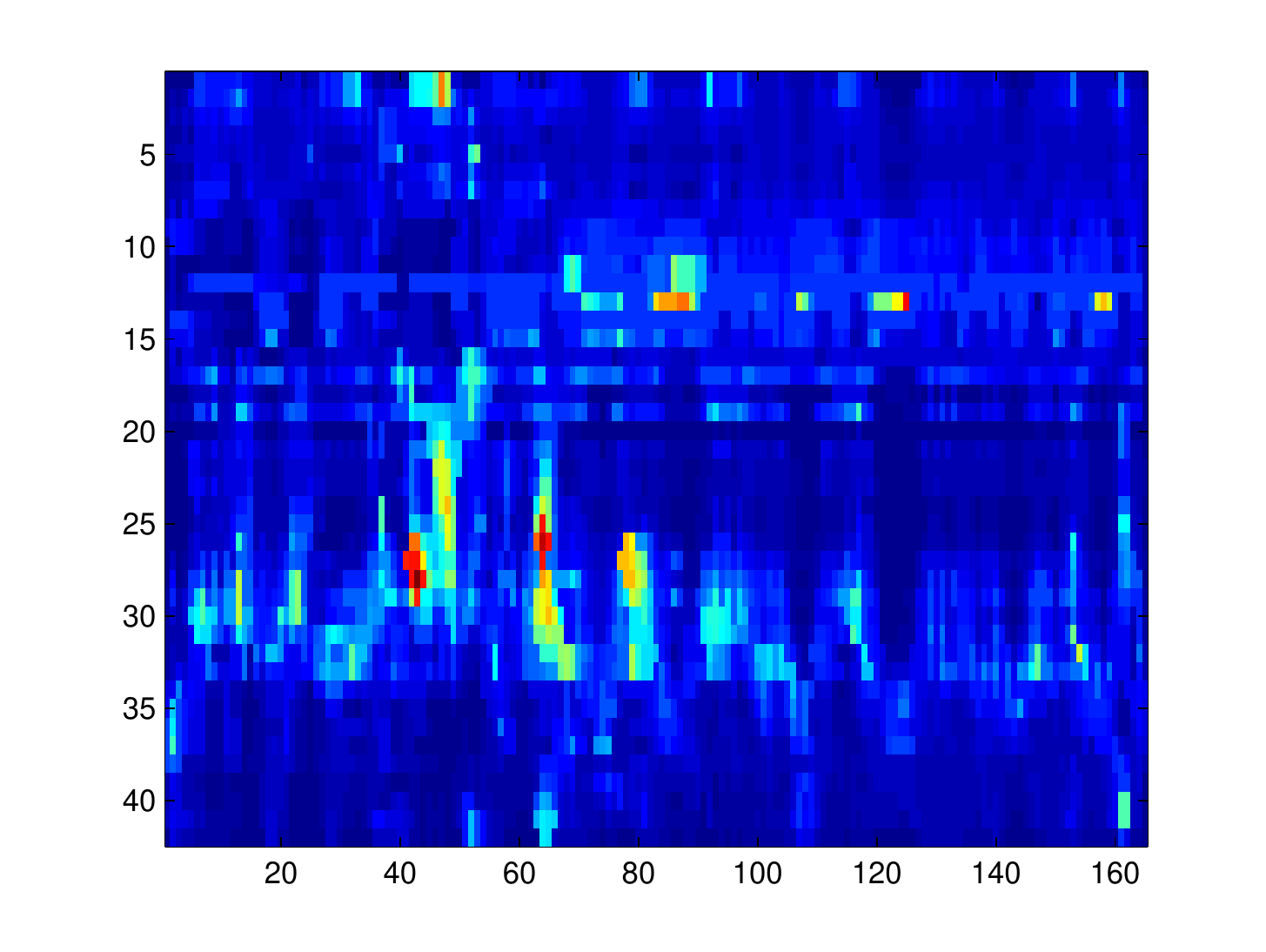}&%
\includegraphics[width=45mm]{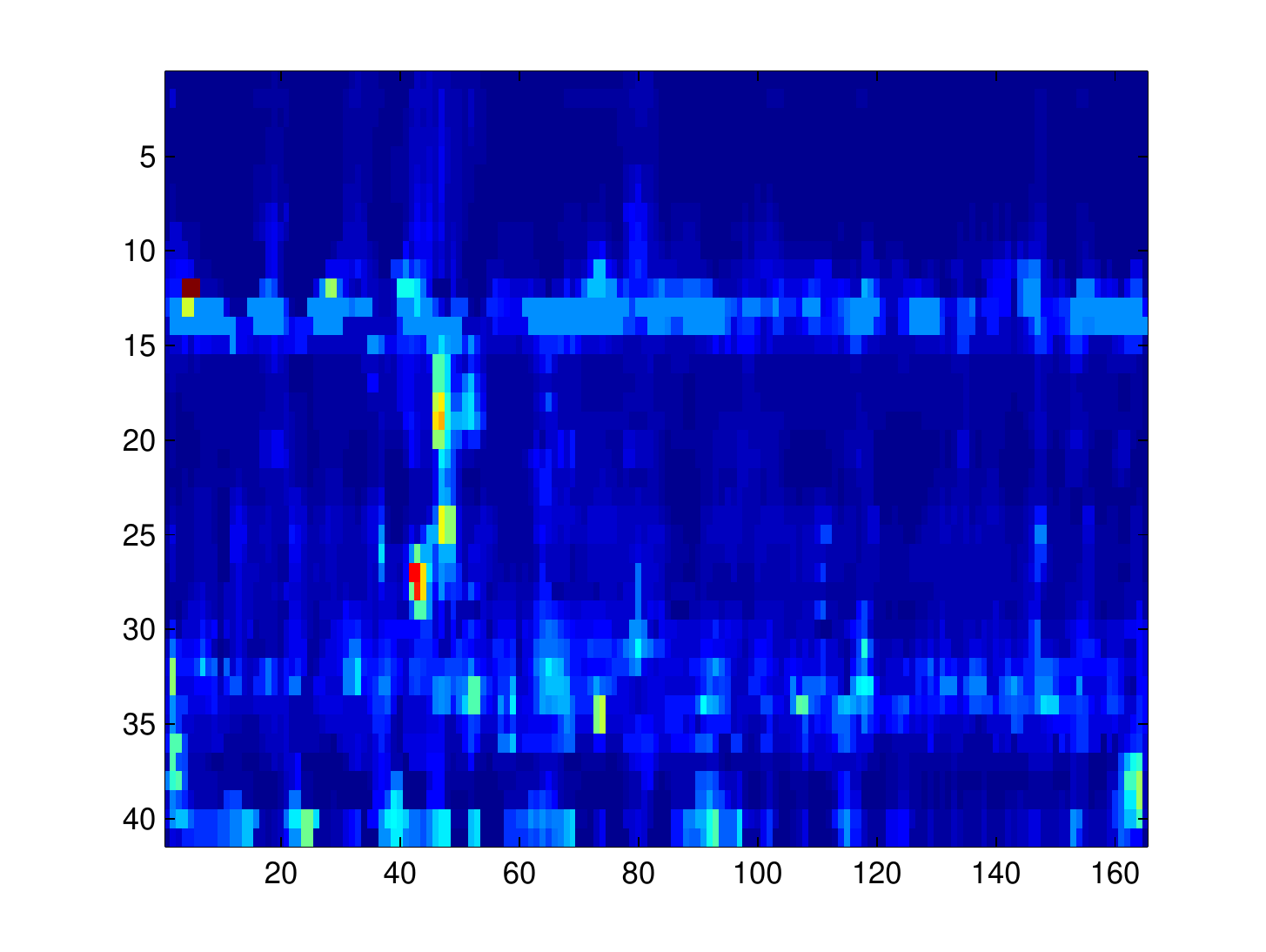} \\
(a)  & (b)  & (c) & (d)\\
\hspace*{-1cm}\includegraphics[width=45mm]{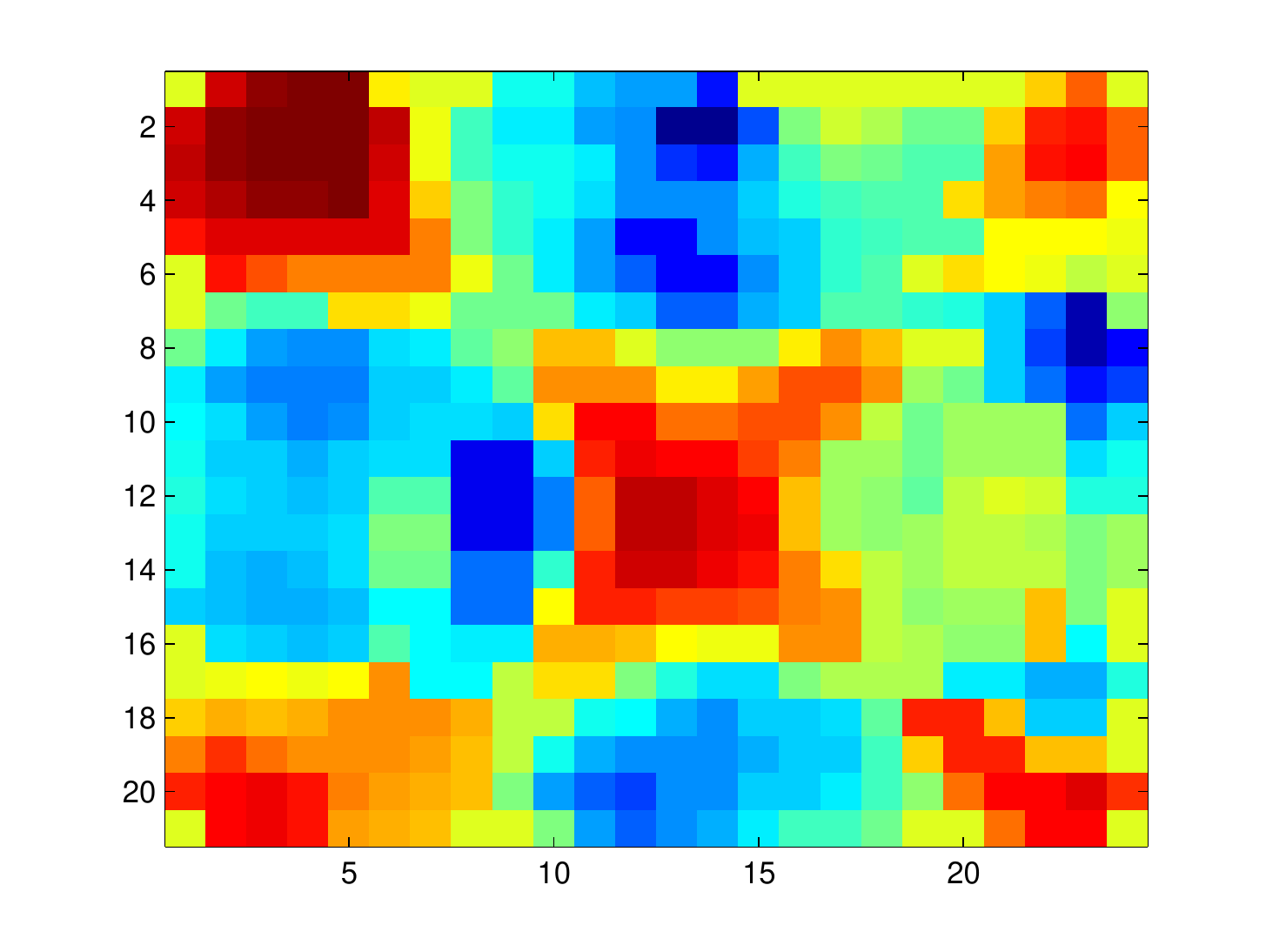} &%
\includegraphics[width=45mm]{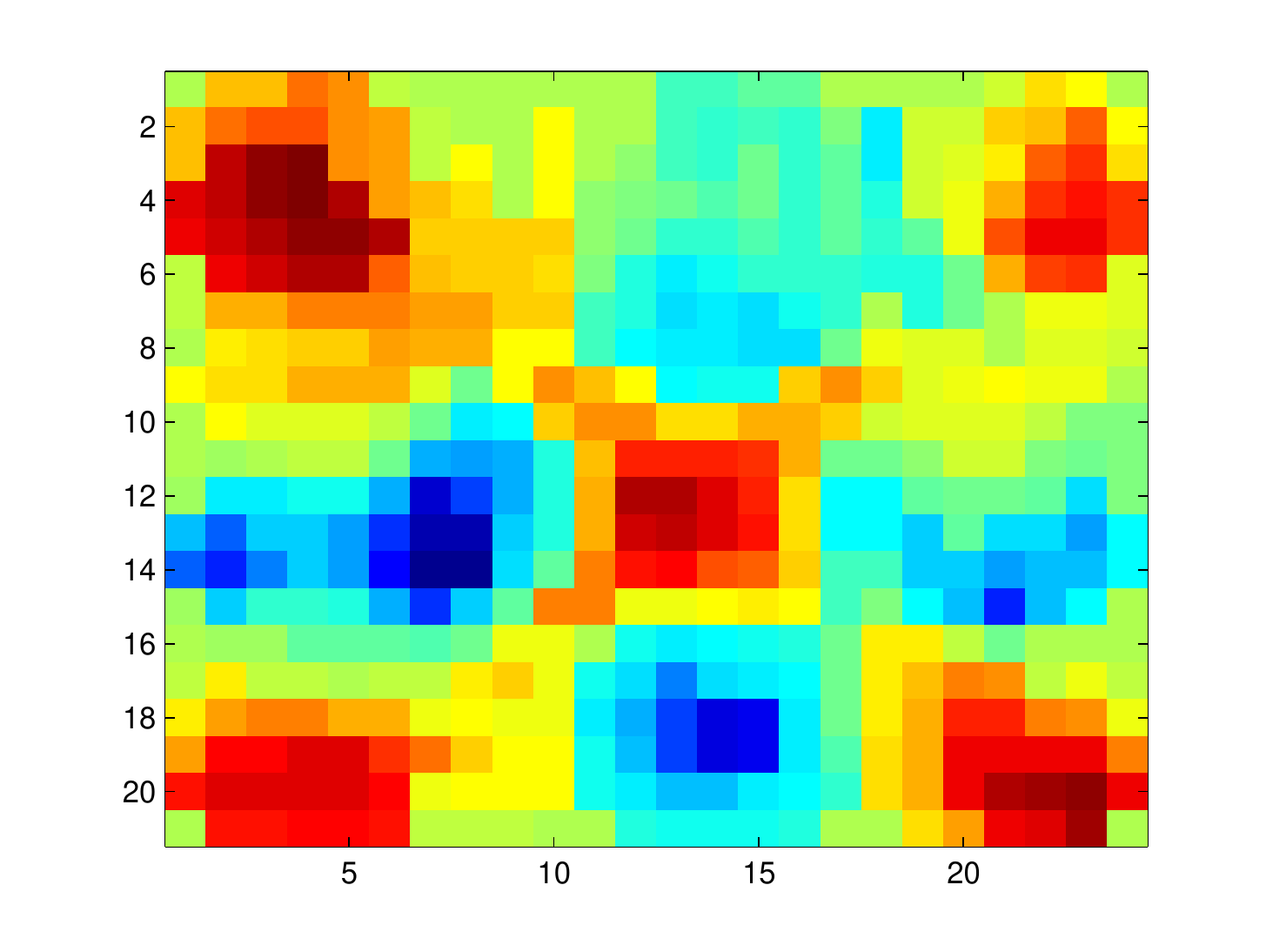} &%
\includegraphics[width=45mm]{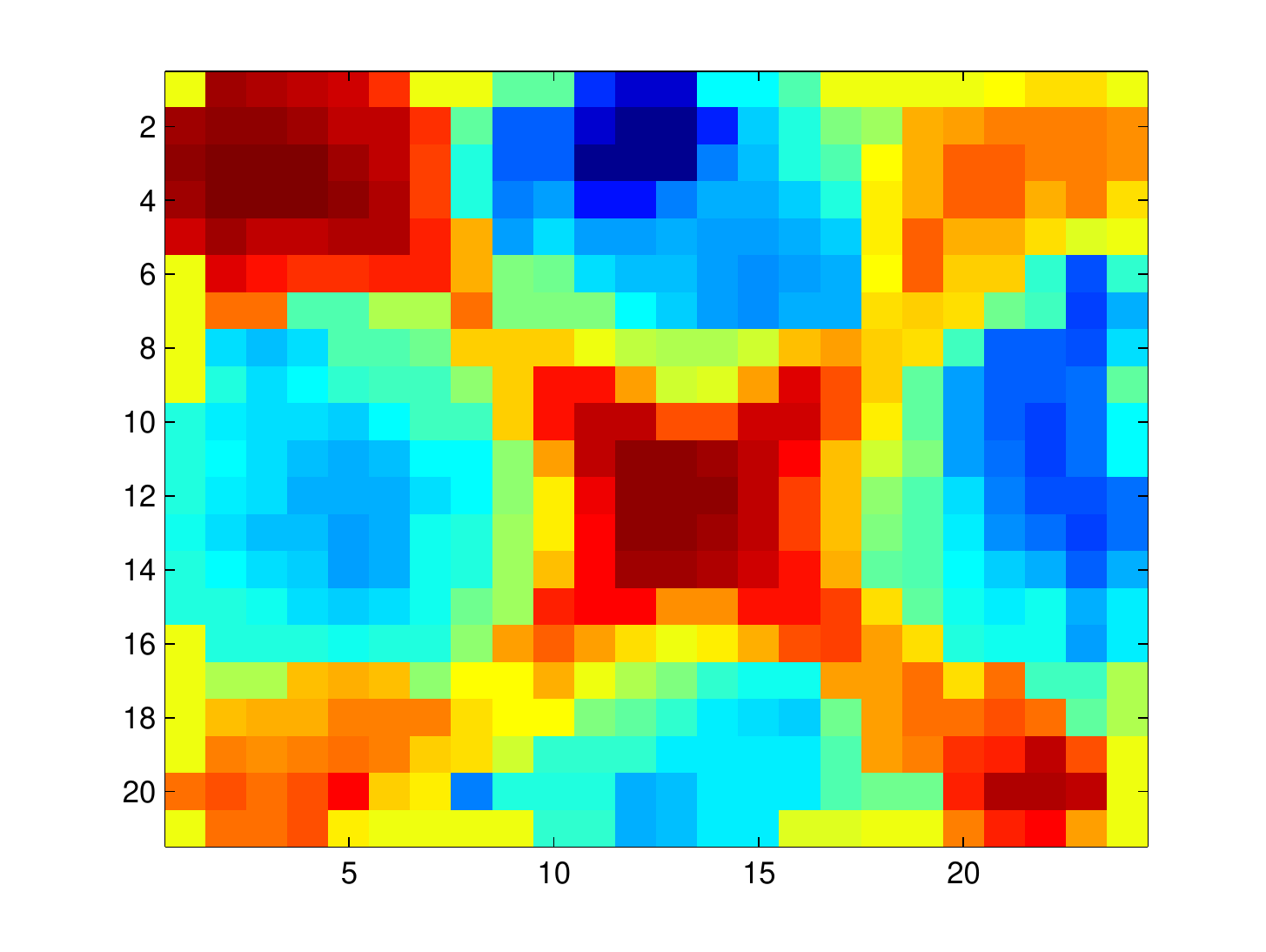}&%
\includegraphics[width=45mm]{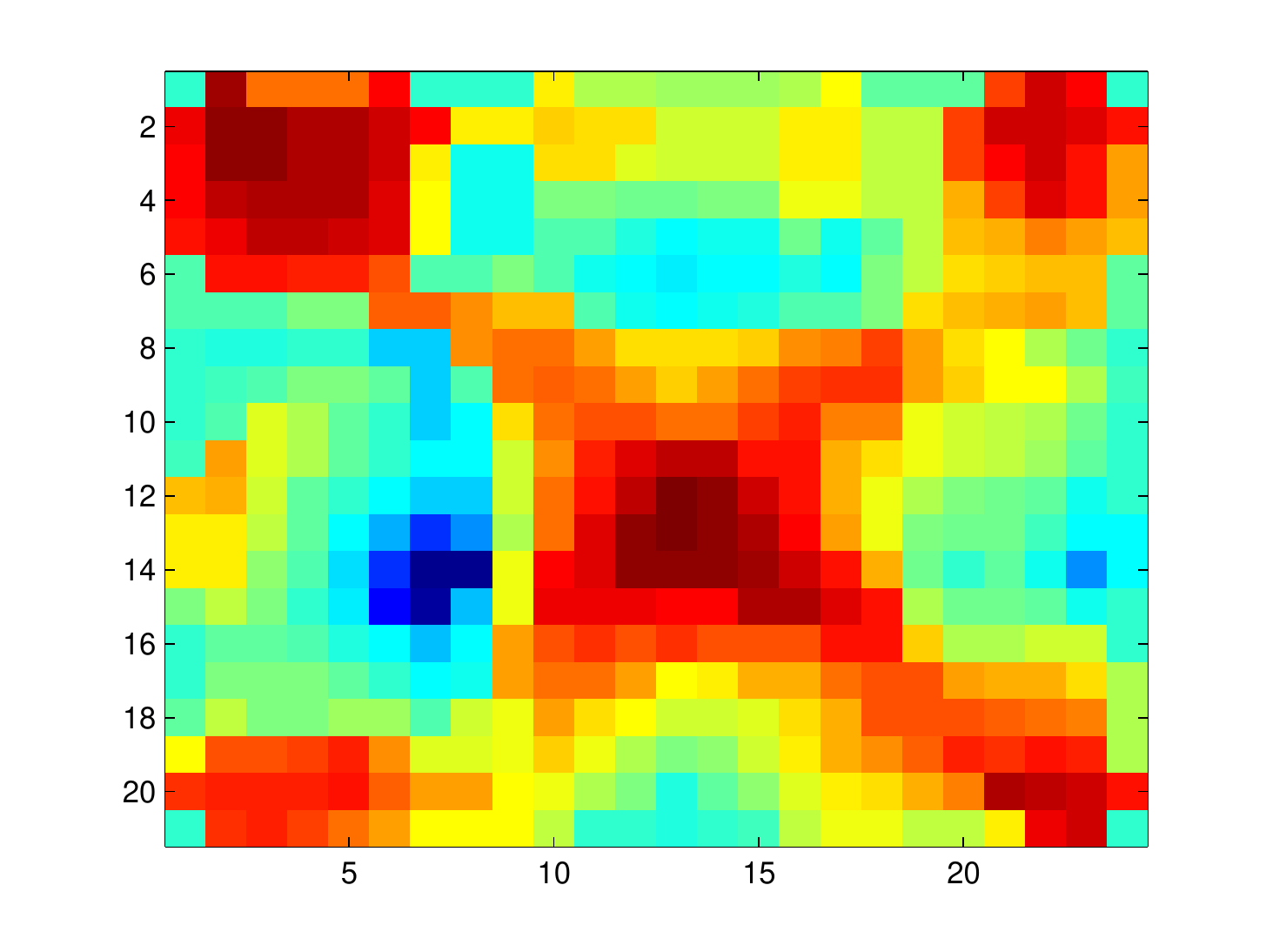} \\
(e)&(f) & (g)  & (h)
\end{tabular}
\caption{TDM and PDM under different viewpoints. (a) - (d) are TDMs and (e) - (h) are PDMs, both corresponding to camera 1 - 4.}
\label{fig:TDMViewInvariance}
\end{figure*}

To study the invariance of TDM and PDM in different viewpoints with different camera parameters, we arbitrarily selected one instance of a subject and its captured videos by 4 different cameras. Similarly we computed the TDM and PDM for each sequence, which are plotted in Figure \ref{fig:TDMViewInvariance}.

As shown in Figure \ref{fig:TDMViewInvariance}, the computed TDM for a subject under various camera setups have similar peaks and valleys. Although minor variations exist, it is still easy to distinguish individuals from the visual differences of TDMs. The same observation is made regarding the PDMs. These observations show that as expected TDM and PDM are view invariant, since they are based on invariants associated with cross-homographies.

\subsection{Gender Recognition}
In this section, we discuss the application of our solution to gender recognition using action style representations TDM and PDM. We are provided with a set of training sequences which are labeled as $w_0$ (female) and $w_1$ (male). Our goal is to find a classifier that correctly categorizes an input action sequence $\mathcal{T}$ as $w_0$ or $w_1$. As discussed in previous sections, TDM and PDM provide good representations of action styles. However, due to the irregularities in human motion, the same action may be performed slightly differently even by the same subject in various instances, producing thus different patterns in TDM and PDM. While some of these patterns are essential for recognizing action styles, others are merely noise, making style analysis extremely challenging. Another challenge of course is the dimensionality of the problem. To tackle these problems, we first serialize the two matrices $\mathbf{T}$ and $\mathbf{P}$ as two vectors $\mathbf{T}_s$ and $\mathbf{P}_s$ and then stacked them together in a single $d$-dimensional vector
\begin{equation}
\mathbf{x} = \left[\mathbf{T}_s, \mathbf{P}_s\right].
\end{equation}
In the next two sections, we propose two frameworks for gender classification using $\mathbf{x}$ that summarizes TDM and PDM.
%
%Through feature extraction, we can project our training data to a lower dimensional space, and then further classify the extra.
%
\subsubsection{Gender Classification using PCA}
In order to reveal the underlying stylistic information in the TDM and PDM, we need to describe the data in a way that makes ``critical'' (significant) and ``trivial'' (insignificant) triplets or pose patterns better discriminated. Principal Component Analysis (PCA) provides a good solution for this purpose, providing both feature selection and dimensionality reduction. Suppose we have a set of $N$ $d$-dimensional samples $\mathbf{x}_1, \dots, \mathbf{x}_N$. The goal here is to find a natural set of $d$ orthonormal basis vectors $\{\mathbf{e}_i\}$ to represent the samples such that the criterion function
\begin{equation}
\mathcal{J}_{d} = \sum_{k=1}^{N}{\left\|\left(\mathbf{m} + \sum_{i=1}^{d}{a^k_i \mathbf{e}_i} - \mathbf{x}_k \right)\right\|^2}
\end{equation}
is minimized, where $\mathbf{m}$ is the sample mean,
\begin{equation}
\mathbf{m} = \frac{1}{n}\sum_{k=1}^{N}\mathbf{x}_k,
\end{equation}
and
\begin{equation}
\mathbf{a}^k = [\begin{array}{cccc}a^k_1 & a^k_2 & \dots &a^k_{d'}\end{array}]^T
\end{equation}
are defined as principal components.
The solution is to compute the eigenvalues and eigenvectors of the scatter matrix $\mathbf{S}$
\begin{equation}
\mathbf{S} = \sum_{k=1}^N{(\mathbf{x}_k - \mathbf{m})(\mathbf{x}_k - \mathbf{m})^T},
\end{equation}
and sort the eigenvalues and eigenvectors according to decreasing eigenvalue. The $d'$ largest eigenvectors are then used as the basis vectors $\{\mathbf{e}_i\}$. Usually $d'$ is much smaller than $d$, which implies that the $d'$ dimensions are inherent
subspaces that govern the samples, while the remaining $d - d'$ dimensions are merely noise. A sample $\mathbf{x}_k$ can now be represented by principal components though projecting onto the $d'$ dimensional subspace as $\tilde{\mathbf{x}}_k$:
\begin{equation}\label{eq:styleReconstruct}
\tilde{\mathbf{x}}_k = \mathbf{A}^T(\mathbf{x}_k - \mathbf{m}),
\end{equation}
where
\begin{equation}
\mathbf{A} = \left[ \begin{array}{cccc}
\mathbf{e}_1 &\mathbf{e}_2 & \dots & \mathbf{e}_{d'}
\end{array}\right].
\end{equation}

The basis vectors computed by PCA are in the direction of the largest variance of the training vectors, and they convey the stylistic elements inherent to the specific motion. We call these bases the ``eigenstyles''. These eigenstyles span a style space of the specific motion. When a sample $\mathbf{x}_k$ is projected onto the style space, its vector $\tilde{\mathbf{x}}_k$ describes the significance of these eigenstyles in the sample. We therefore define $\tilde{\mathbf{x}}_k$ as the stylistic feature of the sample. A style sample/representation can be reconstructed with some error based on the eigenstyles and its stylistic feature from equation (\ref{eq:styleReconstruct}).

We adapted the k-nearest neighbor algorithm to classify sequences represented by the PCA based stylistic feature as follows. Suppose we are provided with a set of stylistic feature vectors $\{\tilde{\mathbf{x}}_k| k = 1, 2, \dots, n\}$ after PCA, and their corresponding labels $\left\{\mathcal{L}_k| k=1, 2, \dots, n, \mathbf{L}_k \in \{w_0, w_1\}\right\}$. A target sequence $\mathcal{T}$ is classified as $w_0$ or $w_1$ based on the following procedure:
\begin{enumerate}
\item The PDM and TDM of sequence $\mathcal{T}$ are first computed, and then seialized as a $d$-dimensional vector $\mathbf{x}$.
\item $\mathbf{x}$ is projected onto the eigenstyle space as $\tilde{\mathbf{x}}$.
\item The Euclidian distances between $\tilde{\mathbf{x}}$ and all $\{\tilde{\mathbf{x}}_k$ in the training set are computed, and $\mathcal{T}$ is classified as $w_i$ which is most frequent among the $k$ training vectors nearest to $\tilde{\mathbf{x}}$, where $k$ is the closest odd integer to $\sqrt{n}$.
\end{enumerate}

\subsubsection{Gender Classification using LDA}
The PCA method finds eigenstyles to describe as much deviation in data as possible, and provides good features to describe the data. However, $d-d'$ dimensions that are thrown away in PCA may still contain useful information for our classification task. On the other hand, PCA is an unsupervised technique that seeks features which are efficient for describing data. However, it does not make use of the label information in data. Unlike PCA, Linear Discriminant Analysis (LDA) seeks features that are efficient to discriminate the classes given the labeled data. Suppose the data $\{\mathbf{x}_i| i=1\dots n\}$ are categorized into $w_0$ and $w_1$, LDA projects the data $\mathbf{x}_i$ onto point $y$ on a line $\mathbf{w}$ by a linear combination of the components of $\mathbf{x}$:
\begin{equation}
y = \mathbf{w}^T\mathbf{x},
\end{equation}
and seeks an optimal $\mathbf{w}$ that results in best separation between points with different labels. It is solved by maximizing the objective function:
\begin{equation}
J(\mathbf{w}) = \frac{\mathbf{w}^T\mathbf{S}_B \mathbf{w}}{\mathbf{w}^T\mathbf{S}_W \mathbf{w}},
\end{equation}
where the intra-class scatter matrix $\mathbf{S}_W$ is defined as
\begin{equation}
\mathbf{S}_w = \sum_{\mathbf{x} \in w_0}{(\mathbf{x} - \mathbf{m}_0)(\mathbf{x} - \mathbf{m}_0)^T} +  \sum_{\mathbf{x} \in w_1}{(\mathbf{x} - \mathbf{m}_1)(\mathbf{x} - \mathbf{m}_1)^T},
\end{equation}
the inter-class scatter matrix  $\mathbf{S}_B$ is defined as
\begin{equation}
\mathbf{S}_B = (\mathbf{m}_1 - \mathbf{m}_2)(\mathbf{m}_1 - \mathbf{m}_2)^T,
\end{equation}
and
\begin{equation}
\mathbf{m}_i = \frac{1}{n}\sum_{\mathbf{x}_i \in w_i}{\mathbf{x}_i}.
\end{equation}
As discussed in \cite{duda2001pattern}, $J(\cdot)$ is independent of $\|\mathbf{w}\|$, and the solution of $\mathbf{w}$ that minimizes $J(\cdot)$ is
\begin{equation}
\mathbf{w} = \mathbf{S}^{-1}_w(\mathbf{m}_1 - \mathbf{m}_2).
\end{equation}

Using $\mathbf{w}$, we project our style vectors $\mathbf{x}$ on a line, and the projected scalar value $y$ is our extracted stylistic feature. We thus convert the $d$-dimensional classification problem to a far more manageable one-dimensional one.  All that remains for our task is to find a threshold that separates the projected points into $w_0$ and $w_1$. Here the decision surface is reduced to a scalar value. We assume that the stylistic vectors of both classes exhibit approximately the same distributions, therefore we choose the separation threshold as
\begin{equation}
c = \mathbf{w}(\frac{\mathbf{m}_1+\mathbf{m}_2}{2}).
\end{equation}

%For our problem, linear LDA is sufficient as shown from experiments in the next section. However, when linear discriminant is not complex enough, LDA can be extended to nonlinear discriminant by the Kernel LDA.

\section{Experiments}
In this section, we present experiments to demonstrate the effectiveness of our proposed gender recognition from human motion, based on the proposed stylistic features. We tested our methods on all the 13 actions from the IXMAS data set. IXMAS dataset consists of 13 everyday actions performed 3 times by 11 actors at arbitrary positions and orientations, and observed by 5 cameras set up at various viewpoints. We assumed that body points were tracked. We then arbitrarily selected one of the subjects as reference for each action, and selected a small number of sequences as a training set, which included 2 female subjects and 2 male subjects, with reference, training and testing sets completely disjoint. For each action, the PCA based method and LDA method were applied to classify the testing sequences as male or female.

%With PCA based method, the number of eigenstyles computed for kicking and walking are based on the threshold $\tau$ in equation \ref{eq:eigenvalueThres}. Figure \ref{fig:eigenstyles} illustrate the computed eigenstyles for kicking and walking when we use $\tau = 90\%$.
\begin{figure*}
\centering
\begin{tabular}{ccc}
\includegraphics[width=55mm]{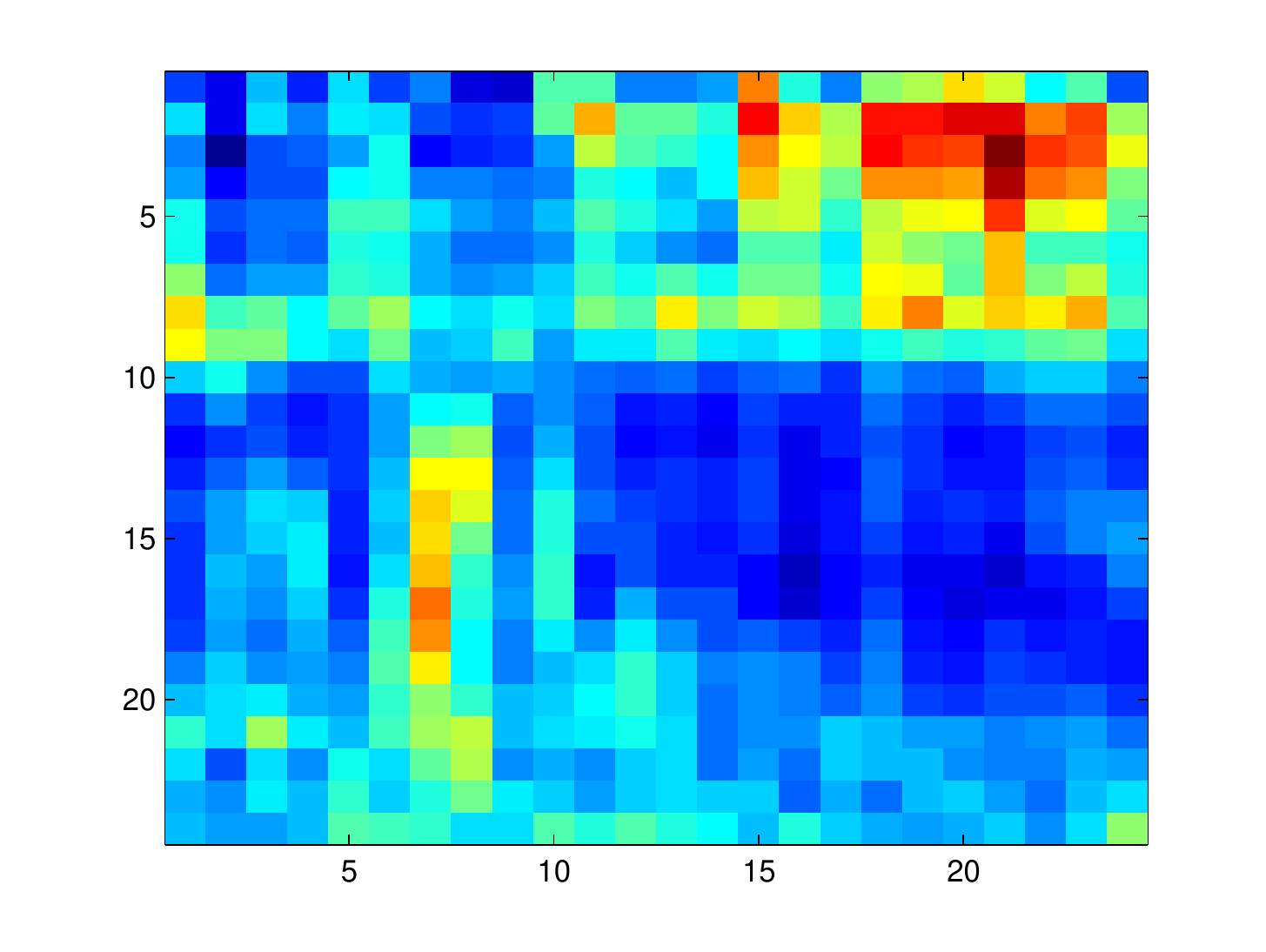} &\includegraphics[width=55mm]{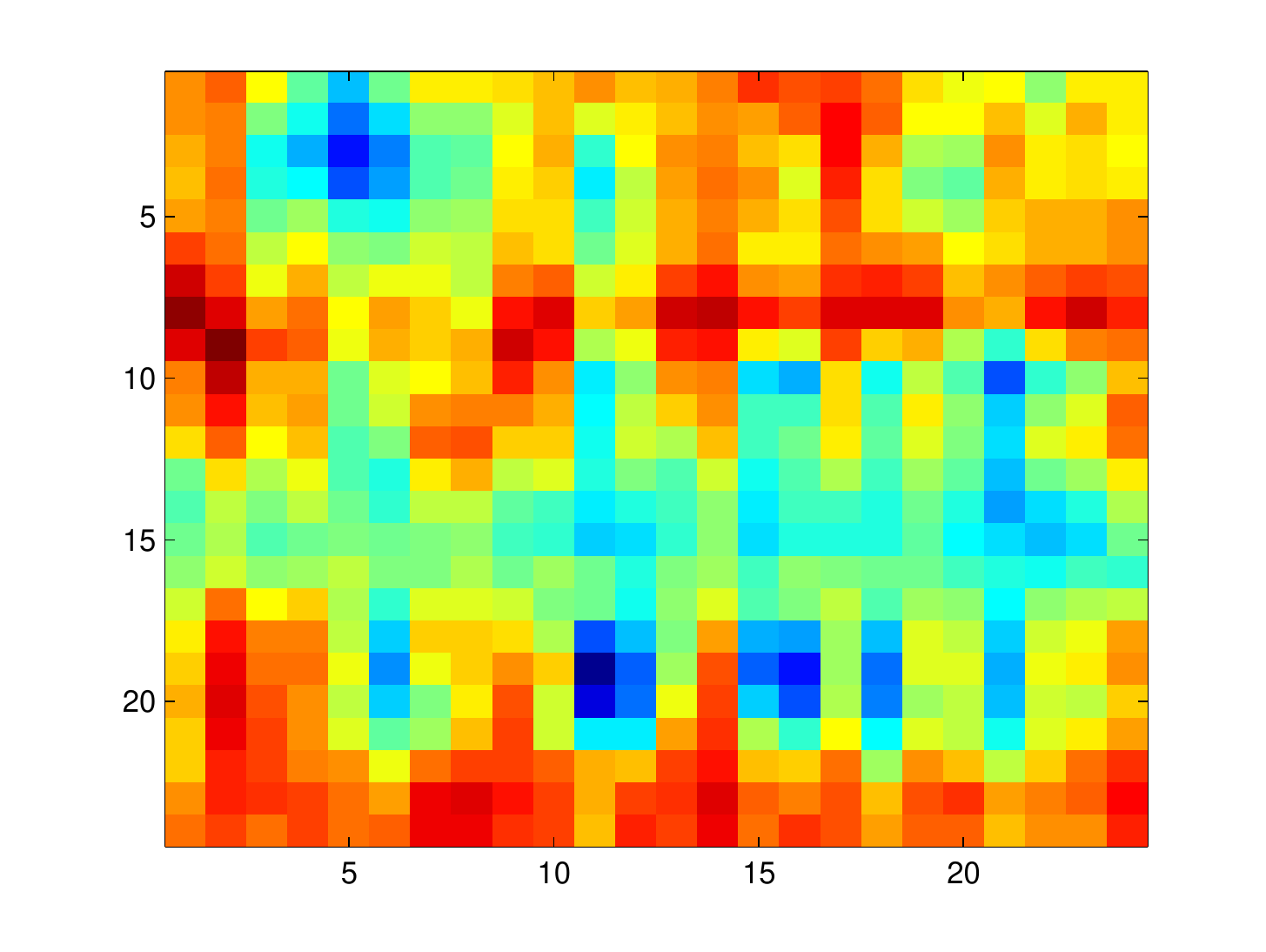} & \includegraphics[width=55mm]{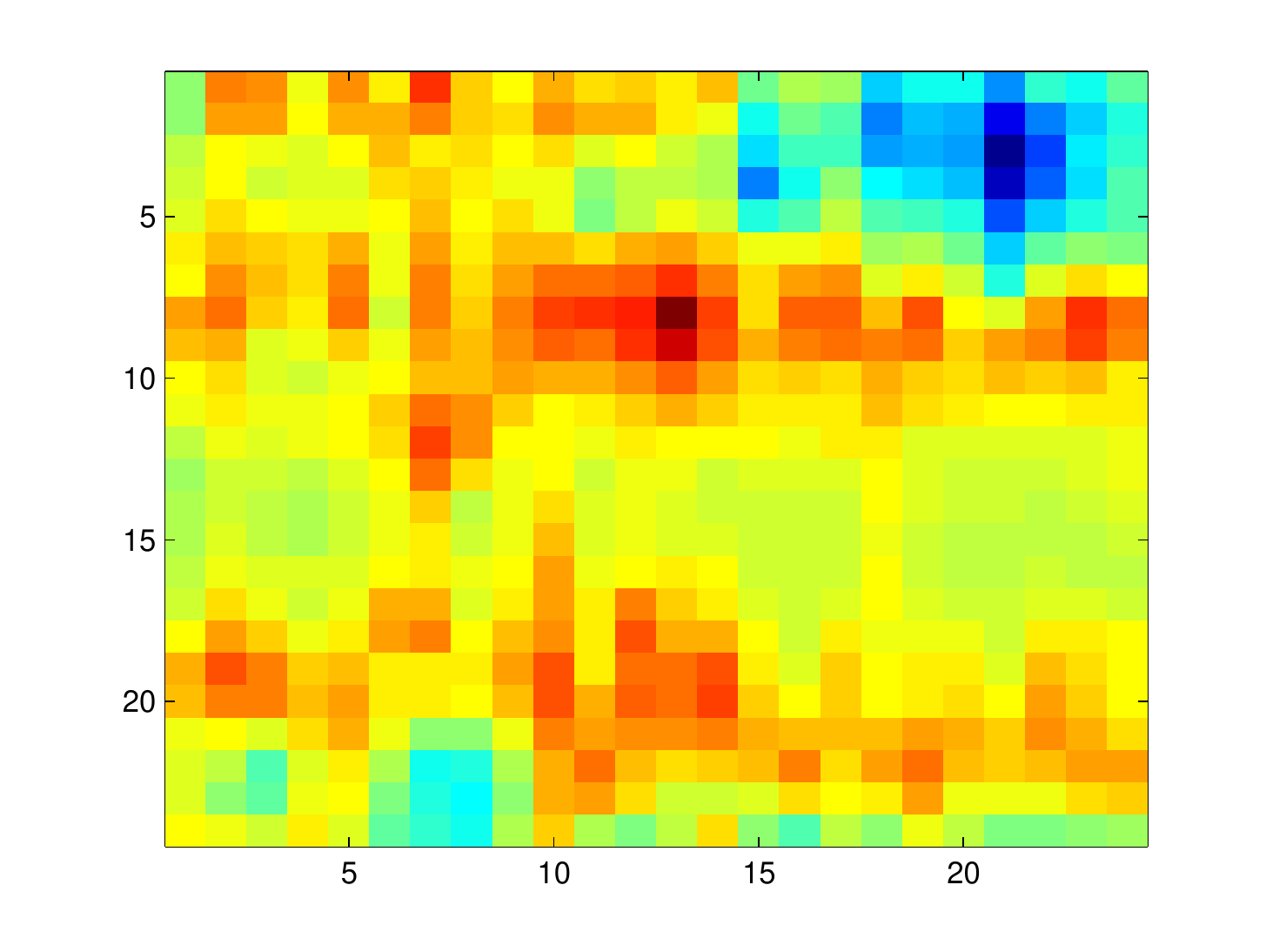} \\
(1)  & (2) & (3) \\
\includegraphics[width=55mm]{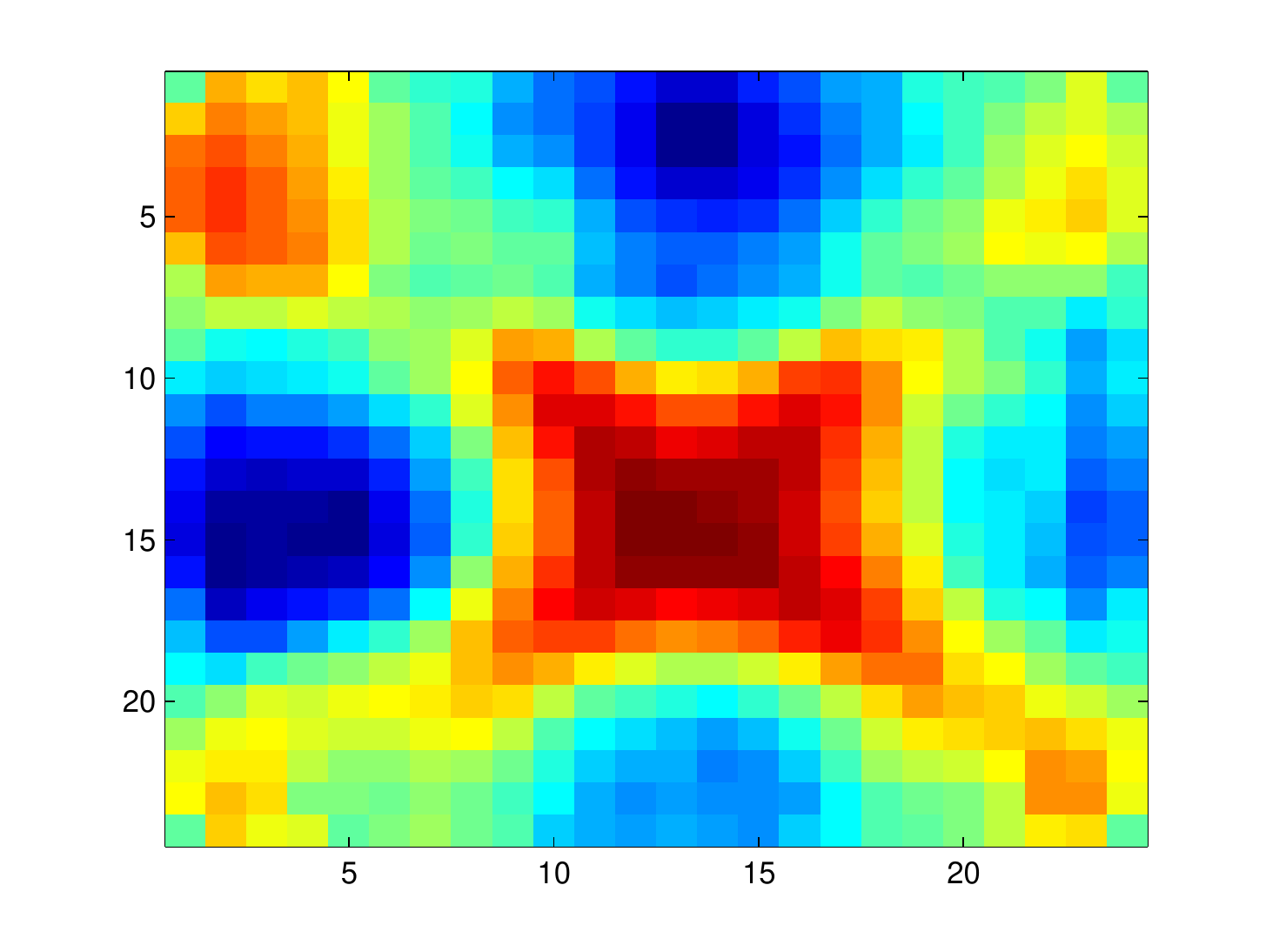} &\includegraphics[width=55mm]{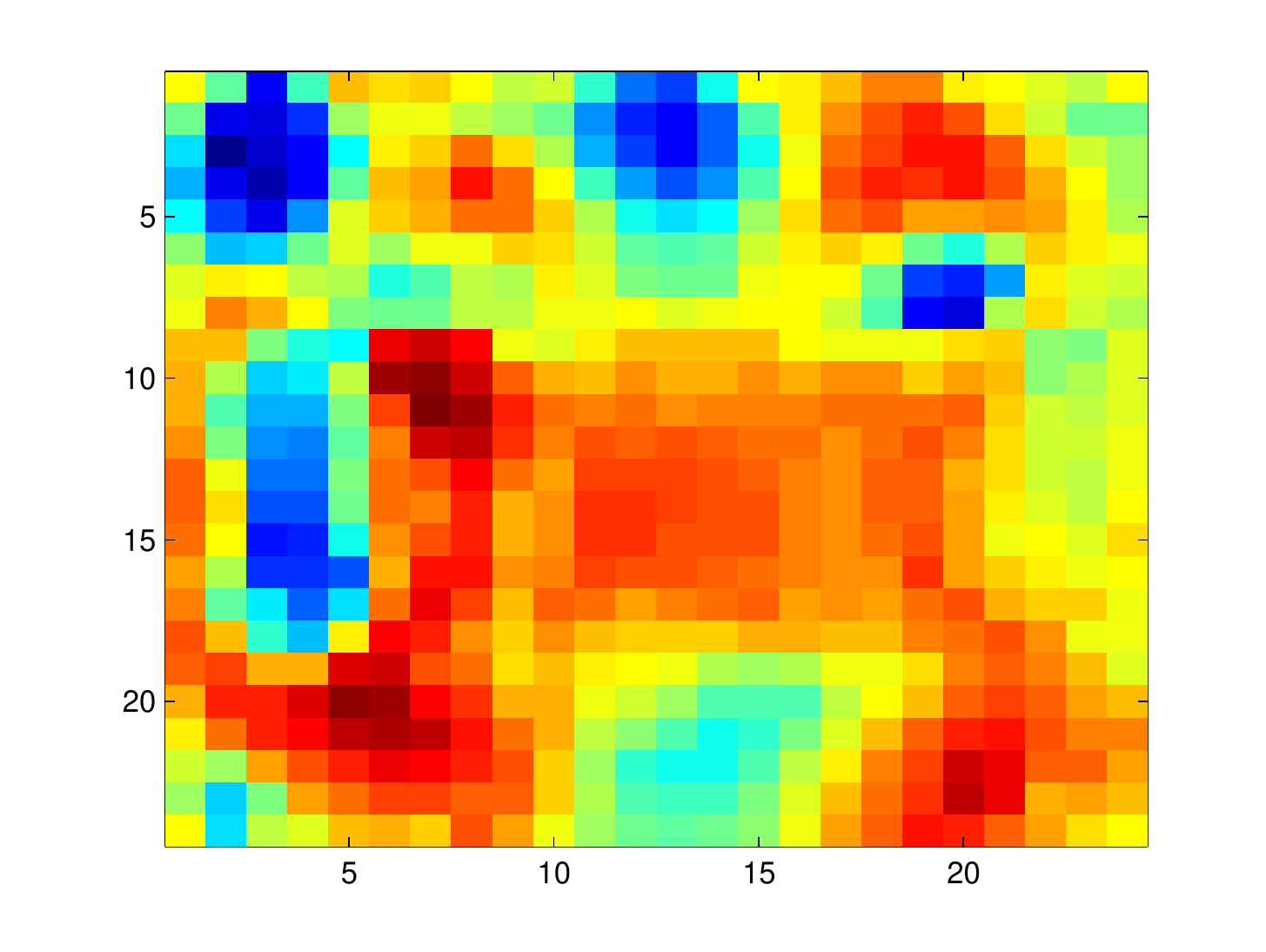} & \includegraphics[width=55mm]{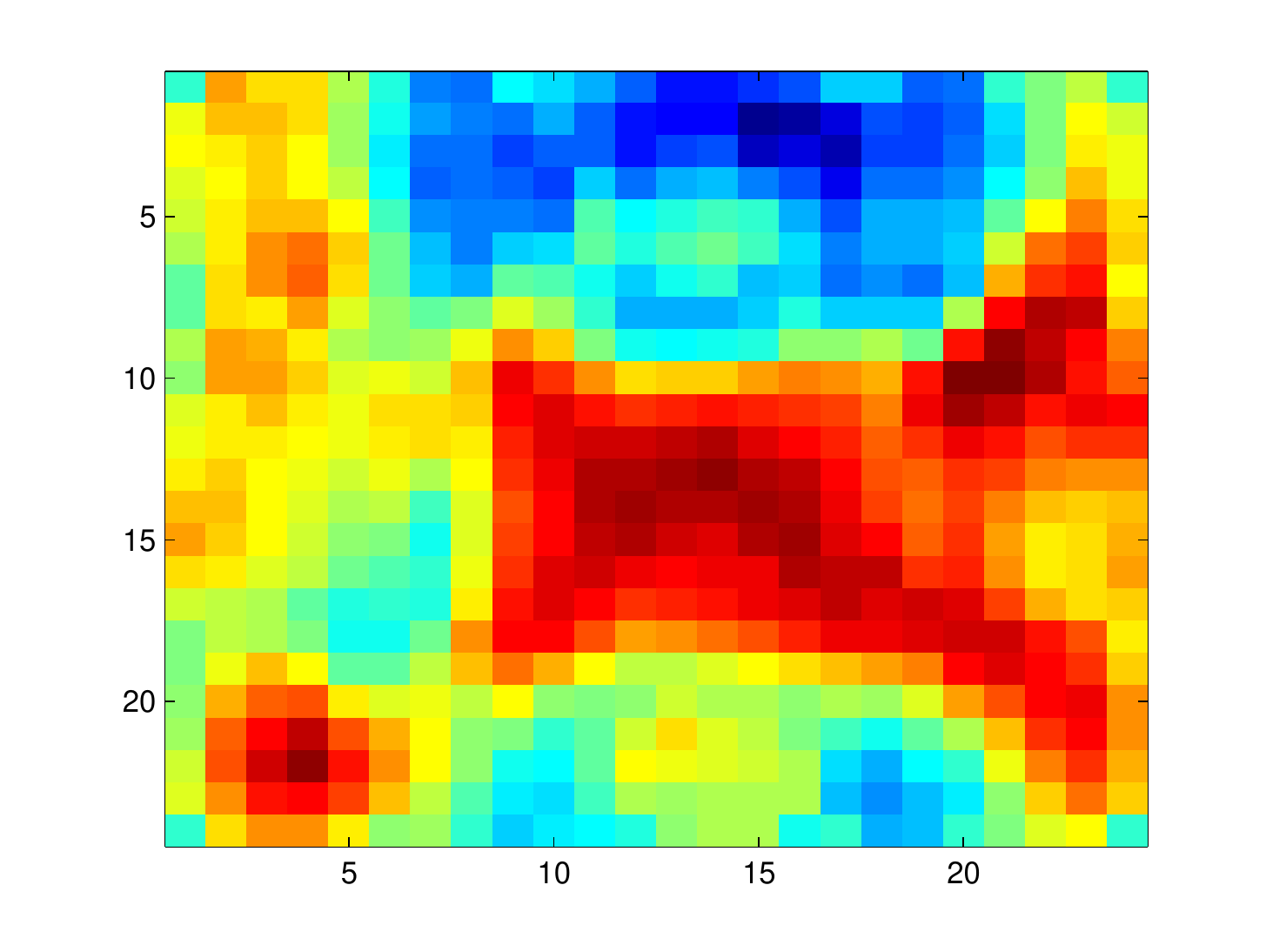} \\
(4) & (5)& (6)
\end{tabular}
\caption{The first 3 eigenstyles computed for kicking action. (1) - (3) are the TDM parts of the eigenstyles, while (4) - (6) are the PDM parts.}
\label{fig:eigenstyles}
\end{figure*}

Figure \ref{fig:eigenstyles} displays the first 3 eigenstyles we obtained for kicking action. We used different values of $d'$ when using our PCA method, and measured the resulted classification rates. Results for 5 of the actions are shown below. We found that the remaining actions in the dataset did not provide sufficient information to distinguish male and female actors. Examples of such actions are "watch time", "cross arms", "wave hand", etc. Our explanation for this is that male and female subjects essentially perform these simple tasks in almost identical manner. Also, these actions are extremely simple and very little parts of the body are involved. As a result, little information is present in the data for gender classification. On the other hand the five actions for which the results are illustrated in the graphs and the tables below involve more sophisticated body part motions, providing thus a better means of distinguishing gender.\\

\begin{figure}[h!]
\centering
\begin{tabular}{cc}
\includegraphics[width=80mm]{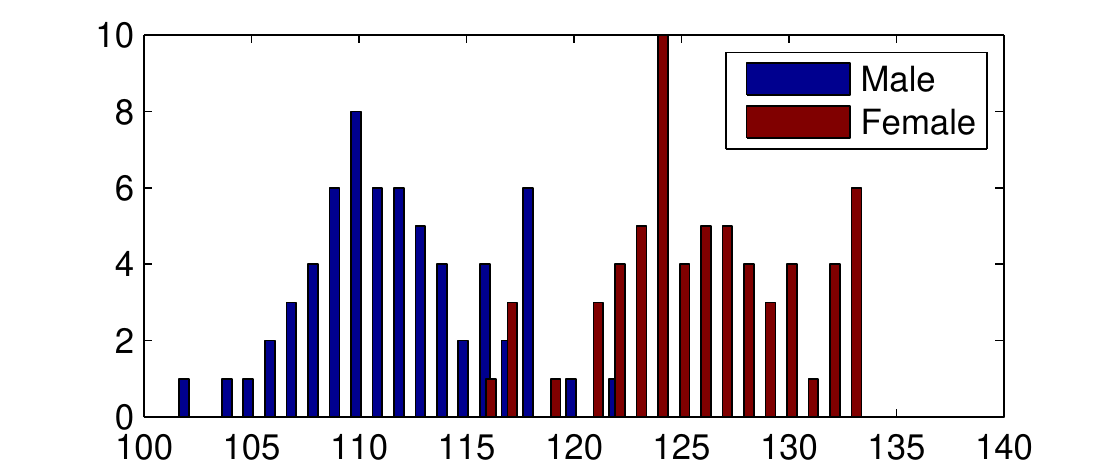} 
\includegraphics[width=80mm]{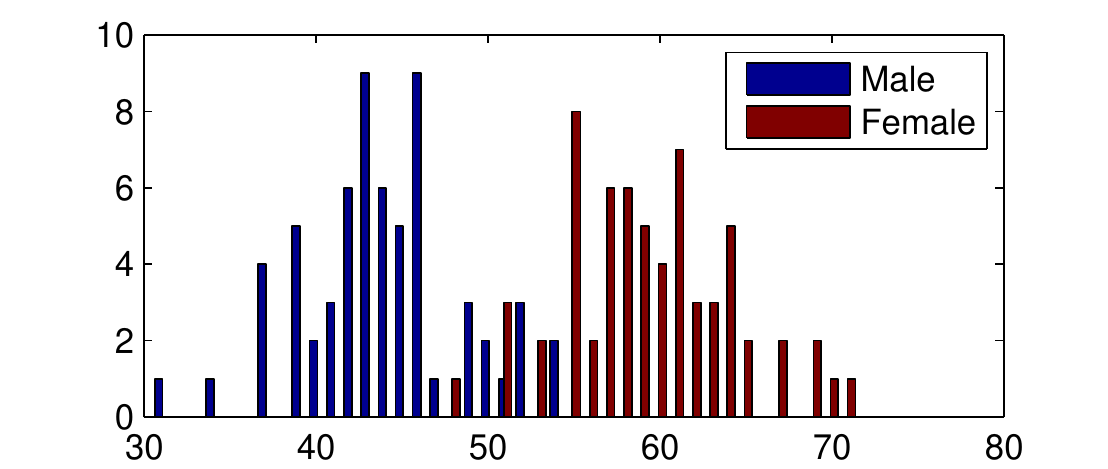}
\end{tabular}
\caption{Distribution of projected points of stylistic vectors. Top is for kicking (threshold: 118.36) and bottom is for walking action (threshold: 72.175).}
\label{fig:recRateFLDA}
\end{figure}

With the LDA method, we projected the high dimensional data onto a one dimension subspace. Figure \ref{fig:recRateFLDA} illustrate examples of the distribution of the projected points for the two actions of ``kicking'' and ``walking''. As shown in the histograms, the data of two classes are well separated, and could be distinguished by simple thresholding. \\

\begin{figure}[h]
\centering
\begin{tabular}{cc}
\includegraphics[width=80mm]{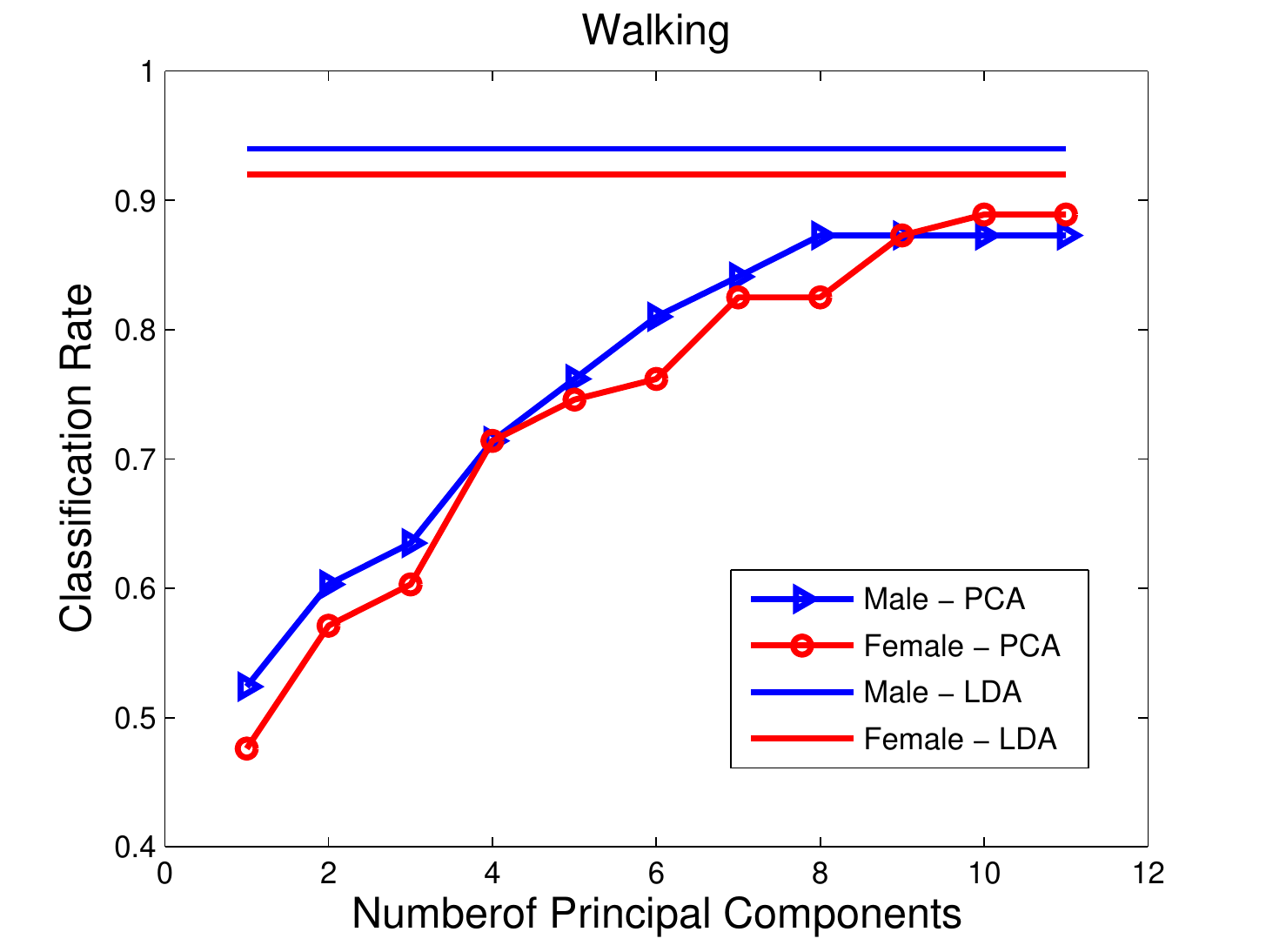} &
\includegraphics[width=80mm]{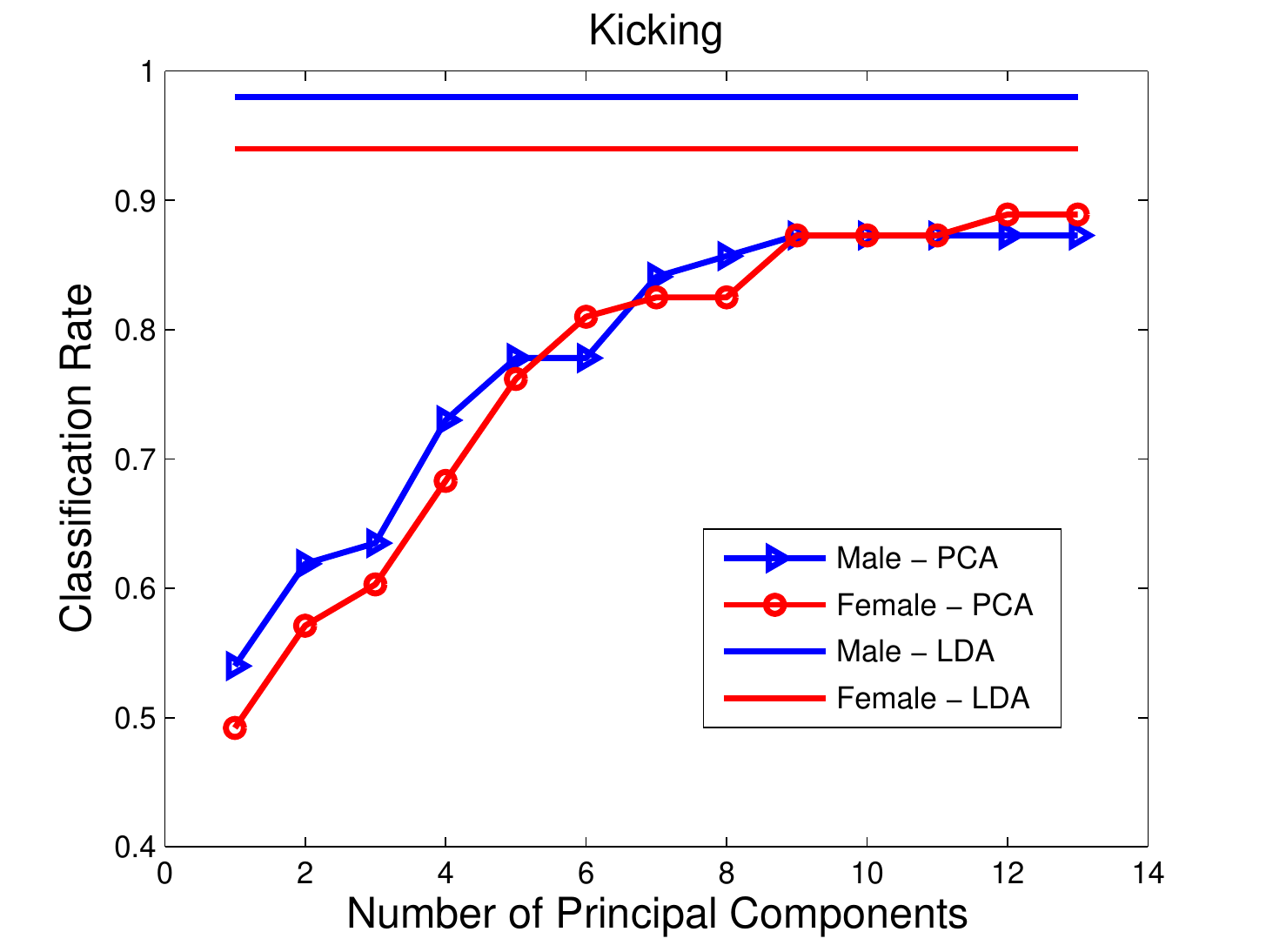} \\
(a)&(b) \\
\includegraphics[width=80mm]{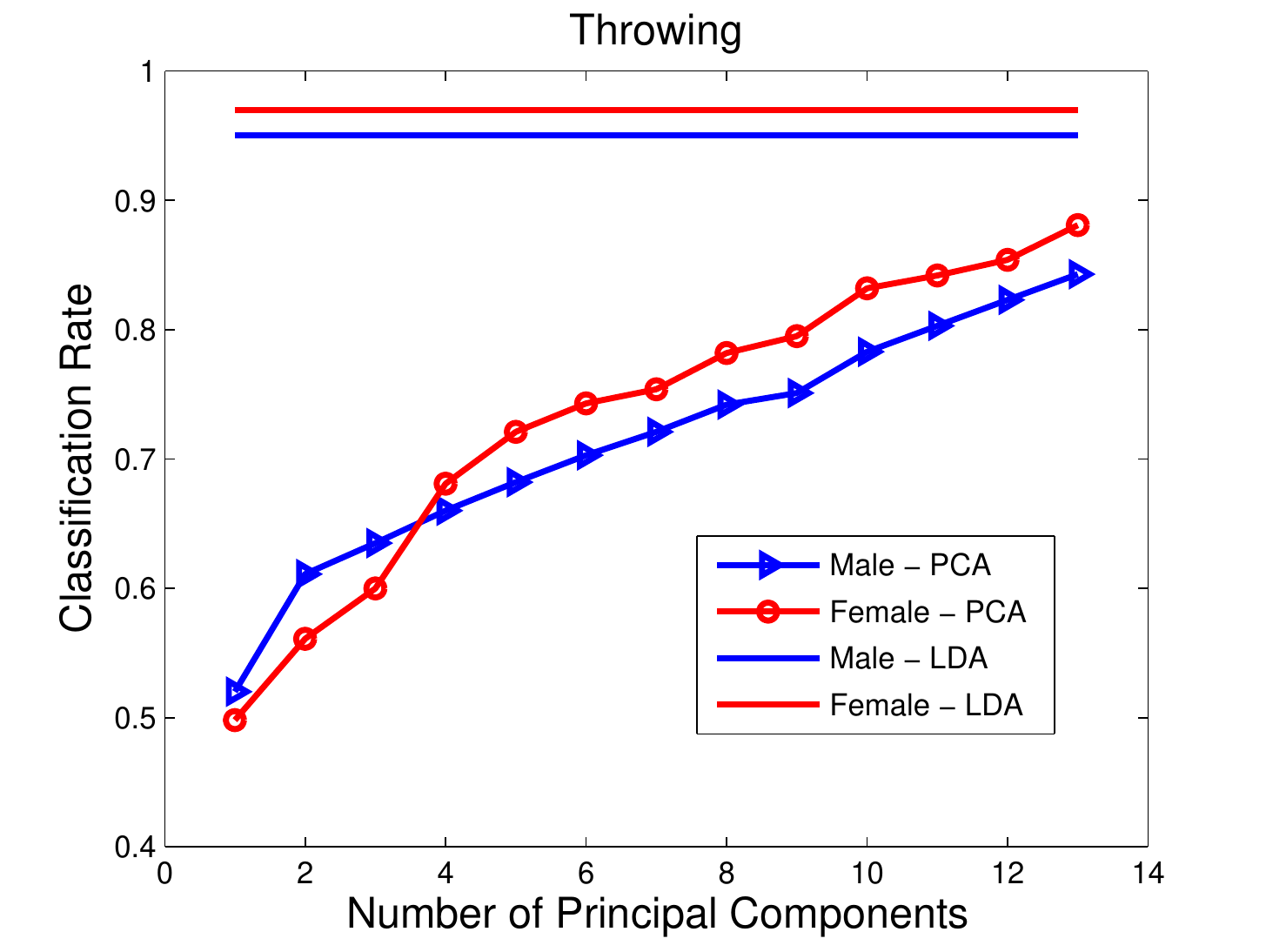} &
\includegraphics[width=80mm]{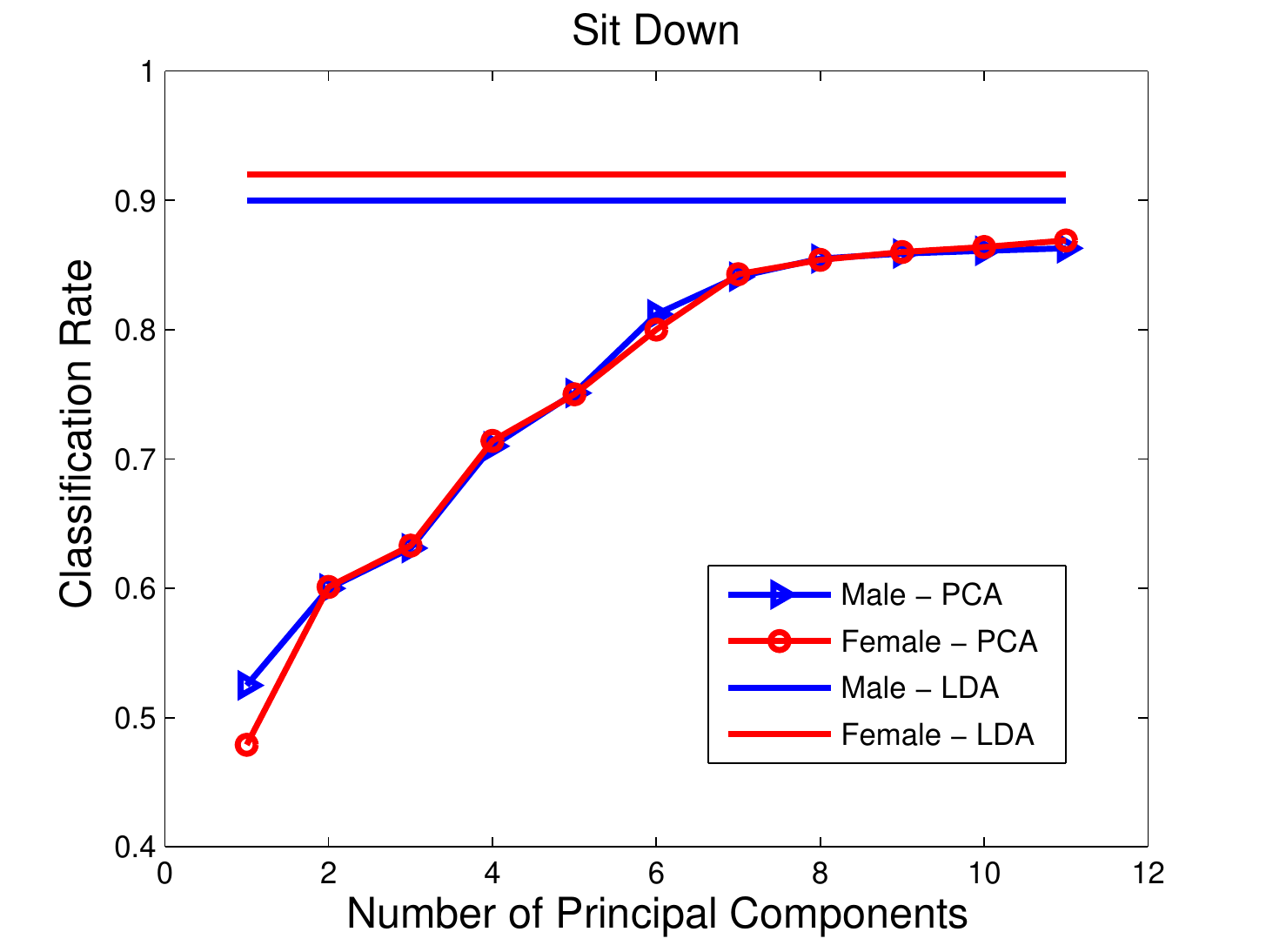} \\
(c)&(d) \\
\end{tabular}
\includegraphics[width=80mm]{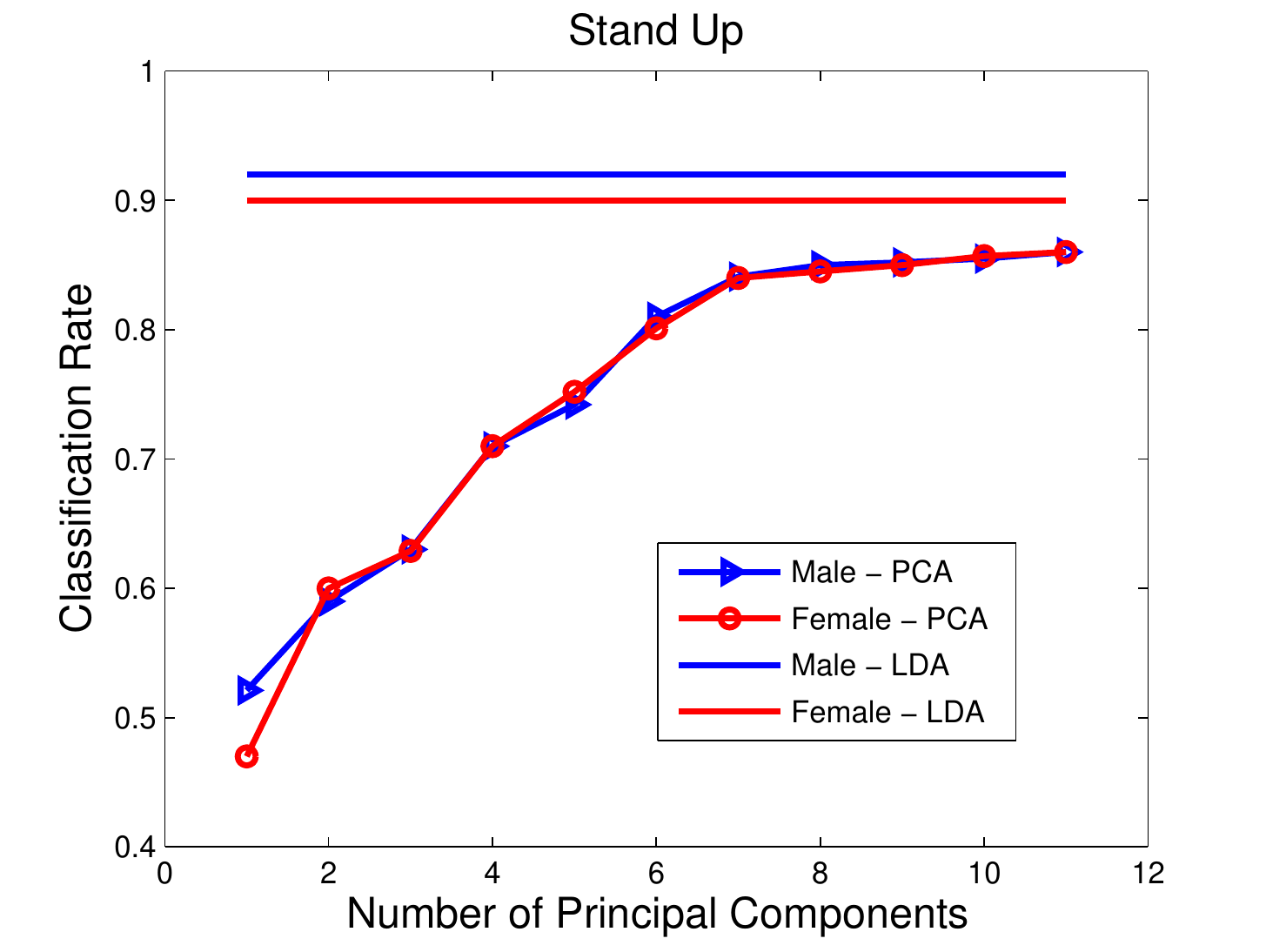}\\
(e)
\caption{Classification rates using PCA method and LDA method. The pictures illustrate the classification rates as a function of $d'$ for ``walking'', ``kicking'', ``throwing'', ``sit down'', and ``stand up'' actions.}
\label{fig:recRatePca1}
\end{figure}

We plot the resulted classification rates based on LDA with those based on PCA in Figures \ref{fig:recRatePca1}: (a)-(e).
As can be seen in these results, the LDA method is more efficient in the task of gender recognition, partially due to the fact that LDA makes better use of labels in the training set, and also the exact features that are more efficient for discriminating classes.

%
%\begin{figure}[htb]
%\centering
%\includegraphics[width=80mm]{golf-golf1}
%\caption{Roles of different triplets in action recognition}
%\label{fig:recRatePca2}
%\end{figure}
\begin{table}[h!]
\caption{Male and female classification rates with different $d'$ for walking action.}
\centering
\begin{tabular}{|l||c@{}|c@{}|c@{}|c@{}|c@{}|c@{}|c@{}|c@{}|c@{}|c@{}|c@{}|c@{}|c@{}|}
\hline
$d'$ &1 &2 &3 &4 &5 &6 &7 &8 &9 &10 &11\\
\hline
Male classification rate & .524 & .603 & .635 & .714 & .762 & .810 & .841 & .873 & .873 & .873 & .873\\
\hline
Female classification rate & .476 & .571 & .603 &.714 & .746 & .762 & .825 & .825 & .873 & .889 & .889\\
\hline
\end{tabular}\label{tab:recRatePca1}
\end{table}

\begin{table}[h!]
\caption{Male and female classification rates with different $d'$ for kicking action.}
\centering
\begin{tabular}{|l||@{}c@{}|@{}c@{}|@{}c@{}|@{}c@{}|@{}c@{}|@{}c@{}|@{}c@{}|@{}c@{}|@{}c@{}|@{}c@{}|@{}c@{}|@{}c@{}|@{}c@{}|@{}c@{}|@{}c@{}|}
%\begin{tabular}{|l||c@{}|c@{}|c@{}|c@{}|c@{}|c@{}|c@{}|c@{}|c@{}|c@{}|c@{}|c@{}|c@{}|}
\hline
 $d'$ &1 &2 &3 &4 &5 &6 &7 &8 &9 &10 &11&12&13\\
\hline
Male classification rate & .540 & .619 & .635 & .730 & .778 & .778 & .841 & .857 & .873 & .873 & .873&.873&.873\\
\hline
Female classification rate & .492 & .571 & .603 &.683 & .762 & .810 & .825 & .825 & .873 & .873 & .873&.889&.889\\
\hline
\end{tabular}\label{tab:recRatePca2}
\end{table}

\begin{table}[h!]
\caption{Male and female classification rates with different $d'$ for throwing action.}
\centering
\begin{tabular}{|l||@{}c@{}|@{}c@{}|@{}c@{}|@{}c@{}|@{}c@{}|@{}c@{}|@{}c@{}|@{}c@{}|@{}c@{}|@{}c@{}|@{}c@{}|@{}c@{}|@{}c@{}|@{}c@{}|@{}c@{}|}
%\begin{tabular}{|l||c@{}|c@{}|c@{}|c@{}|c@{}|c@{}|c@{}|c@{}|c@{}|c@{}|c@{}|c@{}|c@{}|}
\hline
 $d'$ &1 &2 &3 &4 &5 &6 &7 &8 &9 &10 &11&12&13\\
\hline
Male classification rate & 0.520  & 0.611  &  0.635 &   0.660  &  0.682  &  0.703 &   0.721 &   0.742  &  0.751 &   0.783 &   0.803  &  0.823  &  0.843\\
\hline
Female classification rate & 0.498  &  0.561  &  0.600  &  0.681  &  0.721  &  0.743  &  0.754  &  0.782 &  0.795  &  0.832  &  0.842  &  0.854  &  0.8810\\
\hline
\end{tabular}\label{tab:recRatePca3}
\end{table}

\begin{table}[h!]
\caption{Male and female classification rates with different $d'$ for sit down action.}
\centering
\begin{tabular}{|l||c@{}|c@{}|c@{}|c@{}|c@{}|c@{}|c@{}|c@{}|c@{}|c@{}|c@{}|c@{}|c@{}|}
\hline
$d'$ &1 &2 &3 &4 &5 &6 &7 &8 &9 &10 &11\\
\hline
Male classification rate & 0.525  &  0.600  &  0.631  &  0.710  &  0.751  &  0.812  &   0.841  &  0.855 &   0.859  &  0.861 &   0.863\\
\hline
Female classification rate & 0.479  &  0.601 &   0.633  &  0.714  &  0.750  &  0.800  &  0.843  &  0.854 &   0.860  &  0.864 &   0.869\\
\hline
\end{tabular}\label{tab:recRatePca4}
\end{table}

\begin{table}[h!]
\caption{Male and female classification rates with different $d'$ for stand up action.}
\centering
\begin{tabular}{|l||c@{}|c@{}|c@{}|c@{}|c@{}|c@{}|c@{}|c@{}|c@{}|c@{}|c@{}|c@{}|c@{}|}
\hline
$d'$ &1 &2 &3 &4 &5 &6 &7 &8 &9 &10 &11\\
\hline
Male classification rate & 0.521  &  0.590  &  0.630  &  0.710  &  0.742  &  0.810 &   0.841  &  0.850  &  0.852  &  0.855 &   0.860\\
\hline
Female classification rate & 0.470  &  0.600  &  0.629  &  0.710 &   0.752  &  0.801 &   0.840  &  0.845  &  0.850  &  0.857   & 0.8600\\
\hline \\
\end{tabular}\label{tab:recRatePca5}
\end{table}

\clearpage
\section{Conclusion}
We propose two invariant measures that can be used for intra-class classification of actions performed by different subjects that are captured by different cameras from different viewing points. We successfully demonstrate their very powerful property of discriminating action styles by using these measures as the feature vectors within two frameworks based on PCA (eigenstyles) and LDA. Our paper makes several main contributions: (i) our methods are invariant to viewpoint variations and camera parameters due to using view-invariant feature vectors, (ii) very little training set is required for our methods while providing very good performance, (iii) we show with extensive experiments that the proposed new eigenstyles and LDA method can reliably classify genders from video data of different actions. Our results can be readily extended to other applications such as age recognition, human identification using gait, and identification of abnormal action features such as carrying extra weight, or walking on an uneven surface.
% \clearpage
\bibliographystyle{plain}
\bibliography{foroosh,yuping,refs}

\begin{thebibliography}{100}

\bibitem{Ali-Foroosh2015}
Muhamad Ali and Hassan Foroosh.
\newblock Natural scene character recognition without dependency on specific
  features.
\newblock In {\em Proc. International Conference on Computer Vision Theory and
  Applications}, 2015.

\bibitem{Ali-Foroosh2016}
Muhamad Ali and Hassan Foroosh.
\newblock A holistic method to recognize characters in natural scenes.
\newblock In {\em Proc. International Conference on Computer Vision Theory and
  Applications}, 2016.

\bibitem{ali2016character}
Muhammad Ali and Hassan Foroosh.
\newblock Character recognition in natural scene images using rank-1 tensor
  decomposition.
\newblock In {\em Proc. of International Conference on Image Processing
  (ICIP)}, pages 2891--2895, 2016.

\bibitem{alnasser2006image}
Mais Alnasser and Hassan Foroosh.
\newblock Image-based rendering of synthetic diffuse objects in natural scenes.
\newblock In {\em Proc. IAPR Int. Conference on Pattern Recognition}, volume~4,
  pages 787--790, 2006.

\bibitem{Alnasser_Foroosh_rend2006}
Mais Alnasser and Hassan Foroosh.
\newblock Rendering synthetic objects in natural scenes.
\newblock In {\em Proc. of IEEE International Conference on Image Processing
  (ICIP)}, pages 493--496, 2006.

\bibitem{Alnasser_Foroosh_2008}
Mais Alnasser and Hassan Foroosh.
\newblock Phase shifting for non-separable 2d haar wavelets.
\newblock {\em IEEE Transactions on Image Processing}, 16:1061--1068, 2008.

\bibitem{Ashraf_Foroosh_2008}
Nazim Ashraf and Hassan Foroosh.
\newblock Robust auto-calibration of a ptz camera with non-overlapping fov.
\newblock In {\em Proc. International Conference on Pattern Recognition
  (ICPR)}, 2008.

\bibitem{ashraf2012human}
Nazim Ashraf and Hassan Foroosh.
\newblock Human action recognition in video data using invariant characteristic
  vectors.
\newblock In {\em Proc. of IEEE Int. Conf. on Image Processing (ICIP)}, pages
  1385--1388, 2012.

\bibitem{ashraf2015motion}
Nazim Ashraf and Hassan Foroosh.
\newblock Motion retrieval using consistency of epipolar geometry.
\newblock In {\em Proceedings of IEEE International Conference on Image
  Processing (ICIP)}, pages 4219--4223, 2015.

\bibitem{ashraf2007near}
Nazim Ashraf, Imran Junejo, and Hassan Foroosh.
\newblock Near-optimal mosaic selection for rotating and zooming video cameras.
\newblock {\em Proc. of Asian Conf. on Computer Vision}, pages 63--72, 2007.

\bibitem{ashraf2010view}
Nazim Ashraf, Yuping Shen, and Hassan Foroosh.
\newblock View-invariant action recognition using rank constraint.
\newblock In {\em Proc. of IAPR Int. Conf. Pattern Recognition (ICPR)}, pages
  3611--3614, 2010.

\bibitem{Ashraf_etal2012}
Nazim Ashraf, Chuan Sun, and Hassan Foroosh.
\newblock Motion retrieval using low-rank decomposition of fundamental ratios.
\newblock In {\em Proc. IEEE International Conference on Image Processing
  (ICIP)}, pages 1905--1908, 2012.

\bibitem{ashraf2012motion}
Nazim Ashraf, Chuan Sun, and Hassan Foroosh.
\newblock Motion retrival using low-rank decomposition of fundamental ratios.
\newblock In {\em Image Processing (ICIP), 2012 19th IEEE International
  Conference on}, pages 1905--1908, 2012.

\bibitem{Ashraf_etal_2014}
Nazim Ashraf, Chuan Sun, and Hassan Foroosh.
\newblock View-invariant action recognition using projective depth.
\newblock {\em Journal of Computer Vision and Image Understanding (CVIU)},
  123:41--52, 2014.

\bibitem{ashraf2014view}
Nazim Ashraf, Chuan Sun, and Hassan Foroosh.
\newblock View invariant action recognition using projective depth.
\newblock {\em Computer Vision and Image Understanding}, 123:41--52, 2014.

\bibitem{Atalay_Foroosh_2017}
Vildan Atalay and Hassan Foroosh.
\newblock In-band sub-pixel registration of wavelet-encoded images from sparse
  coefficients.
\newblock {\em Signal, Image and Video Processing}, 2017.

\bibitem{Atalay_Foroosh_2017-2}
Vildan~A. Aydin and Hassan Foroosh.
\newblock Motion compensation using critically sampled dwt subbands for
  low-bitrate video coding.
\newblock In {\em Proc. IEEE International Conference on Image Processing
  (ICIP)}, 2017.

\bibitem{balci2006alignment}
Murat Balci, Mais Alnasser, and Hassan Foroosh.
\newblock Alignment of maxillofacial ct scans to stone-cast models using 3d
  symmetry for backscattering artifact reduction.
\newblock In {\em Proceedings of Medical Image Understanding and Analysis
  Conference}, 2006.

\bibitem{balci2006image}
Murat Balci, Mais Alnasser, and Hassan Foroosh.
\newblock Image-based simulation of gaseous material.
\newblock In {\em Proc. of IEEE International Conference on Image Processing
  (ICIP)}, pages 489--492, 2006.

\bibitem{balci2006subpixel}
Murat Balci, Mais Alnasser, and Hassan Foroosh.
\newblock Subpixel alignment of mri data under cartesian and log-polar
  sampling.
\newblock In {\em Proc. of IAPR Int. Conf. Pattern Recognition}, volume~3,
  pages 607--610, 2006.

\bibitem{Balci_Foroosh_phase2005}
Murat Balci and Hassan Foroosh.
\newblock Estimating sub-pixel shifts directly from phase difference.
\newblock In {\em Proc. of IEEE International Conference on Image Processing
  (ICIP)}, pages 1057--1060, 2005.

\bibitem{balci2005estimating}
Murat Balci and Hassan Foroosh.
\newblock Estimating sub-pixel shifts directly from the phase difference.
\newblock In {\em Proc. of IEEE Int. Conf. Image Processing (ICIP)}, volume~1,
  pages I--1057, 2005.

\bibitem{balci2005inferring}
Murat Balci and Hassan Foroosh.
\newblock Inferring motion from the rank constraint of the phase matrix.
\newblock In {\em Proc. IEEE Conf. on Acoustics, Speech, and Signal
  Processing}, volume~2, pages ii--925, 2005.

\bibitem{Balci_Foroosh_metro2005}
Murat Balci and Hassan Foroosh.
\newblock Metrology in uncalibrated images given one vanishing point.
\newblock In {\em Proc. of IEEE International Conference on Image Processing
  (ICIP)}, pages 361--364, 2005.

\bibitem{balci2006real}
Murat Balci and Hassan Foroosh.
\newblock Real-time 3d fire simulation using a spring-mass model.
\newblock In {\em Proc. of Int. Multi-Media Modelling Conference}, pages 8--pp,
  2006.

\bibitem{Balci_Foroosh_2006}
Murat Balci and Hassan Foroosh.
\newblock Sub-pixel estimation of shifts directly in the fourier domain.
\newblock {\em IEEE Trans. on Image Processing}, 15(7):1965--1972, 2006.

\bibitem{Balci_Foroosh_2006_2}
Murat Balci and Hassan Foroosh.
\newblock Sub-pixel registration directly from phase difference.
\newblock {\em Journal of Applied Signal Processing-special issue on
  Super-resolution Imaging}, 2006:1--11, 2006.

\bibitem{barclay1978tas}
CD~Barclay, JE~Cutting, and LT~Kozlowski.
\newblock {Temporal and spatial factors in gait perception that influence
  gender recognition.}
\newblock {\em Percept Psychophys}, 23(2):145--52, 1978.

\bibitem{beardsworth1981aro}
T.~Beardsworth and T.~Buckner.
\newblock {The ability to recognize oneself from a video recording of one's
  movements without seeing one's body}.
\newblock {\em Bulletin of the Psychonomic Society}, 18(1):19--22, 1981.

\bibitem{benabdelkader2002mbr}
C.~BenAbdelkader, R.~Cutler, L.~Davis, et~al.
\newblock {Motion-based recognition of people in eigengait space}.
\newblock {\em International Conference on Automatic Face and Gesture
  Recognition}, pages 267--272, 2002.

\bibitem{berthod1994refining}
M~Berthod, M~Werman, H~Shekarforoush, and J~Zerubia.
\newblock Refining depth and luminance information using super-resolution.
\newblock In {\em Proc. of IEEE Conf. Computer Vision and Pattern Recognition
  (CVPR)}, pages 654--657, 1994.

\bibitem{berthod1994reconstruction}
Marc Berthod, Hassan Shekarforoush, Michael Werman, and Josiane Zerubia.
\newblock Reconstruction of high resolution 3d visual information.
\newblock In {\em IEEE Conf. Computer Vision and Pattern Recognition (CVPR)},
  pages 654--657, 1994.

\bibitem{bhutta2006blind}
Adeel Bhutta and Hassan Foroosh.
\newblock Blind blur estimation using low rank approximation of cepstrum.
\newblock {\em Image Analysis and Recognition}, pages 94--103, 2006.

\bibitem{bhutta2011selective}
Adeel~A Bhutta, Imran~N Junejo, and Hassan Foroosh.
\newblock Selective subtraction when the scene cannot be learned.
\newblock In {\em Proc. of IEEE International Conference on Image Processing
  (ICIP)}, pages 3273--3276, 2011.

\bibitem{bobick2001rhm}
AF~Bobick and JW~Davis.
\newblock {The recognition of human movement using temporal templates}.
\newblock {\em TPAMI}, 23(3):257--267, 2001.

\bibitem{bobick2001gea}
AF~Bobick and A.~Johnson.
\newblock {Gait extraction and description by evidence-gathering}.
\newblock {\em Proc. of the IEEE Conference on Computer Vision and Pattern
  Recognition}, 707, 2001.

\bibitem{boyraz122014action}
Hakan Boyraz, Syed~Zain Masood, Baoyuan Liu, Marshall Tappen, and Hassan
  Foroosh.
\newblock Action recognition by weakly-supervised discriminative region
  localization.

\bibitem{brand2000sm}
M.~Brand and A.~Hertzmann.
\newblock {Style machines}.
\newblock {\em Proceedings of the 27th annual conference on Computer graphics
  and interactive techniques}, pages 183--192, 2000.

\bibitem{Cakmakci_etal2008}
Ozan Cakmakci, Gregory~E. Fasshauer, Hassan Foroosh, Kevin~P. Thompson, and
  Jannick~P. Rolland.
\newblock Meshfree approximation methods for free-form surface representation
  in optical design with applications to head-worn displays.
\newblock In {\em Proc. SPIE Conf. on Novel Optical Systems Design and
  Optimization XI}, volume 7061, 2008.

\bibitem{Cakmakci_etal_2008_2}
Ozan Cakmakci, Brendan Moore, Hassan Foroosh, and Jannick Rolland.
\newblock Optimal local shape description for rotationally non-symmetric
  optical surface design and analysis.
\newblock {\em Optics Express}, 16(3):1583--1589, 2008.

\bibitem{Cakmakci_etal_2008}
Ozan Cakmakci, Sophie Vo, Hassan Foroosh, and Jannick Rolland.
\newblock Application of radial basis functions to shape description in a
  dual-element off-axis magnifier.
\newblock {\em Optics Letters}, 33(11):1237--1239, 2008.

\bibitem{chi2000eme}
D.~Chi, M.~Costa, L.~Zhao, and N.~Badler.
\newblock {The EMOTE model for effort and shape}.
\newblock {\em Proceedings of the 27th annual conference on Computer graphics
  and interactive techniques}, pages 173--182, 2000.

\bibitem{collins2002sbh}
R.~Collins, R.~Gross, and J.~Shi.
\newblock {Silhouette-based human identification from body shape and gait}.
\newblock {\em 5th Intl. Conf. on Automatic Face and Gesture Recognition},
  2002.

\bibitem{damkjer2014mesh}
Kristian~L Damkjer and Hassan Foroosh.
\newblock Mesh-free sparse representation of multidimensional {LIDAR} data.
\newblock In {\em Proc. of International Conference on Image Processing
  (ICIP)}, pages 4682--4686, 2014.

\bibitem{davis2002aar}
J.~Davis and S.~Taylor.
\newblock {Analysis and Recognition of Walking Movements}.
\newblock {\em INTERNATIONAL CONFERENCE ON PATTERN RECOGNITION}, 16:315--318,
  2002.

\bibitem{davis2001vcc}
J.W. Davis.
\newblock {Visual Categorization of Children and Adult Walking Styles}.
\newblock {\em LECTURE NOTES IN COMPUTER SCIENCE}, pages 295--300, 2001.

\bibitem{davis2004etm}
J.W. Davis and H.~Gao.
\newblock {An expressive three-mode principal components model for gender
  recognition}.
\newblock {\em Journal of Vision}, 4(5):362--377, 2004.

\bibitem{davis2002efm}
J.W. Davis and V.S. Kannappan.
\newblock {Expressive features for movement exaggeration}.
\newblock {\em International Conference on Computer Graphics and Interactive
  Techniques}, pages 182--182, 2002.

\bibitem{duda2001pattern}
R.O. Duda, P.E. Hart, and D.G. Stork.
\newblock {\em {Pattern classification}}.
\newblock Citeseer, 2001.

\bibitem{Einsele_Foroosh_2015}
Farshideh Einsele and Hassan Foroosh.
\newblock Recognition of grocery products in images captured by cellular
  phones.
\newblock In {\em Proc. International Conference on Computer Vision and Image
  Processing (ICCVIP)}, 2015.

\bibitem{foroosh2003adaptive}
H~Foroosh.
\newblock Adaptive estimation of motion using generalized cross validation.
\newblock In {\em 3rd International (IEEE) Workshop on Statistical and
  Computational Theories of Vision}, 2003.

\bibitem{foroosh2001closed}
Hassan Foroosh.
\newblock A closed-form solution for optical flow by imposing temporal
  constraints.
\newblock In {\em Proc. of IEEE International Conf. on Image Processing
  (ICIP)}, volume~3, pages 656--659, 2001.

\bibitem{foroosh2004adaptive}
Hassan Foroosh.
\newblock An adaptive scheme for estimating motion.
\newblock In {\em Proc. of IEEE International Conf. on Image Processing
  (ICIP)}, volume~3, pages 1831--1834, 2004.

\bibitem{Foroosh_2005}
Hassan Foroosh.
\newblock Pixelwise adaptive dense optical flow assuming non-stationary
  statistics.
\newblock {\em IEEE Trans. on Image Processing}, 14(2):222--230, 2005.

\bibitem{foroosh2004sub}
Hassan Foroosh and Murat Balci.
\newblock Sub-pixel registration and estimation of local shifts directly in the
  fourier domain.
\newblock In {\em Proc. International Conference on Image Processing (ICIP)},
  volume~3, pages 1915--1918, 2004.

\bibitem{Foroosh_Balci_2004}
Hassan Foroosh and Murat Balci.
\newblock Subpixel registration and estimation of local shifts directly in the
  fourier domain.
\newblock In {\em Proc. of IEEE International Conference on Image Processing
  (ICIP)}, volume~3, pages 1915--1918, 2004.

\bibitem{Foroosh_etal_2002}
Hassan Foroosh, Josiane Zerubia, and Marc Berthod.
\newblock Extension of phase correlation to subpixel registration.
\newblock {\em IEEE Trans. on Image Processing}, 11(3):188--200, 2002.

\bibitem{fu2004expression}
Tao Fu and Hassan Foroosh.
\newblock Expression morphing from distant viewpoints.
\newblock In {\em Proc. of IEEE International Conference on Image Processing
  (ICIP)}, volume~5, pages 3519--3522, 2004.

\bibitem{huang1999rhg}
PS~Huang, CJ~Harris, and MS~Nixon.
\newblock {Recognising humans by gait via parametric canonical space}.
\newblock {\em Artificial Intelligence in Engineering}, 13(4):359--366, 1999.

\bibitem{jain2008super}
Apurva Jain, Supraja Murali, Nicolene Papp, Kevin Thompson, Kye-sung Lee,
  Panomsak Meemon, Hassan Foroosh, and Jannick~P Rolland.
\newblock Super-resolution imaging combining the design of an optical coherence
  microscope objective with liquid-lens based dynamic focusing capability and
  computational methods.
\newblock In {\em Optical Engineering \& Applications}, pages 70610C--70610C.
  International Society for Optics and Photonics, 2008.

\bibitem{johansson1973vpb}
G.~JOHANSSON.
\newblock {Visual perception of biological motion and a model for its
  analysis}.
\newblock {\em Perception and Psychophysics}, 14:201--211, 1973.

\bibitem{junejo1dynamic}
I~Junejo, A~Bhutta, and Hassan Foroosh.
\newblock Dynamic scene modeling for object detection using single-class svm.
\newblock In {\em Proc. of IEEE International Conference on Image Processing
  (ICIP)}, volume~1, pages 1541--1544, 2010.

\bibitem{junejo2008cross}
I.~Junejo, E.~Dexter, I.~Laptev, and P.~Perez.
\newblock {Cross-view action recognition from temporal self-similarities}.
\newblock In {\em European Conference on Computer Vision}, volume~12, 2008.

\bibitem{junejo2006dissecting}
Imran Junejo and Hassan Foroosh.
\newblock Dissecting the image of the absolute conic.
\newblock In {\em Proc. of IEEE Int. Conf. on Video and Signal Based
  Surveillance}, pages 77--77, 2006.

\bibitem{junejo2006robust}
Imran Junejo and Hassan Foroosh.
\newblock Robust auto-calibration from pedestrians.
\newblock In {\em Proc. IEEE International Conference on Video and Signal Based
  Surveillance}, pages 92--92, 2006.

\bibitem{Junejo_Foroosh2007-2}
Imran Junejo and Hassan Foroosh.
\newblock Euclidean path modeling from ground and aerial views.
\newblock In {\em Proc. International Conference on Computer Vision (ICCV)},
  pages 1--7, 2007.

\bibitem{Junejo_Foroosh2007-1}
Imran Junejo and Hassan Foroosh.
\newblock Trajectory rectification and path modeling for surveillance.
\newblock In {\em Proc. International Conference on Computer Vision (ICCV)},
  pages 1--7, 2007.

\bibitem{Junejo_Foroosh2007-3}
Imran Junejo and Hassan Foroosh.
\newblock Using calibrated camera for euclidean path modeling.
\newblock In {\em Proceedings of IEEE International Conference on Image
  Processing (ICIP)}, pages 205--208, 2007.

\bibitem{Junejo_Foroosh_2008}
Imran Junejo and Hassan Foroosh.
\newblock Euclidean path modeling for video surveillance.
\newblock {\em Image and Vision Computing (IVC)}, 26(4):512--528, 2008.

\bibitem{Junejo_Foroosh_2010}
Imran Junejo and Hassan Foroosh.
\newblock Camera calibration and geo-location estimation from two shadow
  trajectories.
\newblock {\em Computer Vision and Image Understanding (CVIU)}, 114:915--927,
  2010.

\bibitem{Junejo_etal_2010}
Imran Junejo and Hassan Foroosh.
\newblock Gps coordinates estimation and camera calibration from solar shadows.
\newblock {\em Computer Vision and Image Understanding (CVIU)},
  114(9):991--1003, 2010.

\bibitem{Junejo_etal_2011}
Imran Junejo and Hassan Foroosh.
\newblock Optimizing ptz camera calibration from two images.
\newblock {\em Machine Vision and Applications (MVA)}, pages 1--15, 2011.

\bibitem{junejo2007robust}
Imran~N Junejo, Nazim Ashraf, Yuping Shen, and Hassan Foroosh.
\newblock Robust auto-calibration using fundamental matrices induced by
  pedestrians.
\newblock In {\em Proc. International Conf. on Image Processing (ICIP)},
  volume~3, pages III--201, 2007.

\bibitem{Junejo_etal_2013}
Imran~N. Junejo, Adeel Bhutta, and Hassan Foroosh.
\newblock Single-class svm for dynamic scene modeling.
\newblock {\em Signal Image and Video Processing}, 7(1):45--52, 2013.

\bibitem{junejo2007trajectory}
Imran~N. Junejo and Hassan Foroosh.
\newblock Trajectory rectification and path modeling for video surveillance.
\newblock In {\em Proc. International Conference on Computer Vision (ICCV)},
  pages 1--7, 2007.

\bibitem{Junejo_Foroosh_solar2008}
Imran~N. Junejo and Hassan Foroosh.
\newblock Estimating geo-temporal location of stationary cameras using shadow
  trajectories.
\newblock In {\em Proc. European Conference on Computer Vision (ECCV)}, 2008.

\bibitem{Junejo_Foroosh_GPS2008}
Imran~N. Junejo and Hassan Foroosh.
\newblock Gps coordinate estimation from calibrated cameras.
\newblock In {\em Proc. International Conference on Pattern Recognition
  (ICPR)}, 2008.

\bibitem{junejo2008gps}
Imran~N Junejo and Hassan Foroosh.
\newblock Gps coordinate estimation from calibrated cameras.
\newblock In {\em Proc. International Conference on Pattern Recognition
  (ICPR)}, pages 1--4, 2008.

\bibitem{Junejo_Foroosh_Givens2008}
Imran~N. Junejo and Hassan Foroosh.
\newblock Practical ptz camera calibration using givens rotations.
\newblock In {\em Proc. IEEE International Conference on Image Processing
  (ICIP)}, 2008.

\bibitem{Junejo_Foroosh_calib2008}
Imran~N. Junejo and Hassan Foroosh.
\newblock Practical pure pan and pure tilt camera calibration.
\newblock In {\em Proc. International Conference on Pattern Recognition
  (ICPR)}, 2008.

\bibitem{Junejo_Foroosh_PTZ2008}
Imran~N. Junejo and Hassan Foroosh.
\newblock Refining ptz camera calibration.
\newblock In {\em Proc. International Conference on Pattern Recognition
  (ICPR)}, 2008.

\bibitem{Junejo_Foroosh_SolCalib2008}
Imran~N. Junejo and Hassan Foroosh.
\newblock Using solar shadow trajectories for camera calibration.
\newblock In {\em Proc. IEEE International Conference on Image Processing
  (ICIP)}, 2008.

\bibitem{kale2003tvi}
A.~Kale, A.K.R. Chowdhury, and R.~Chellappa.
\newblock {Towards a view invariant gait recognition algorithm}.
\newblock {\em Proceedings of IEEE Conference on Advanced Video and Signal
  Based Surveillance}, pages 143--150, 2003.

\bibitem{kale2003gah}
A.~Kale, N.~Cuntoor, B.~Yegnanarayana, AN~Rajagopalan, and R.~Chellappa.
\newblock {Gait Analysis for Human Identification}.
\newblock {\em LECTURE NOTES IN COMPUTER SCIENCE}, pages 706--714, 2003.

\bibitem{lee2003lpm}
L.~Lee, G.~Dalley, and K.~Tieu.
\newblock {Learning pedestrian models for silhouette refinement}.
\newblock {\em International Conference on Computer Vision}, 2003.

\bibitem{lee2002gar}
L.~Lee and WEL Grimson.
\newblock {Gait analysis for recognition and classification}.
\newblock {\em Automatic Face and Gesture Recognition, 2002. Proceedings. Fifth
  IEEE International Conference on}, pages 148--155, 2002.

\bibitem{little1998rpt}
J.~Little and J.~Boyd.
\newblock {Recognizing people by their gait: the shape of motion}.
\newblock {\em Videre: Journal of Computer Vision Research}, 1(2):1--32, 1998.

\bibitem{liu2015sparse}
Baoyuan Liu, Min Wang, Hassan Foroosh, Marshall Tappen, and Marianna Pensky.
\newblock Sparse convolutional neural networks.
\newblock In {\em Proceedings of the IEEE Conference on Computer Vision and
  Pattern Recognition (CVPR)}, pages 806--814, 2015.

\bibitem{lorette1997super}
Anne Lorette, Hassan Shekarforoush, and Josiane Zerubia.
\newblock Super-resolution with adaptive regularization.
\newblock In {\em Proc. International Conf. on Image Processing (ICIP)},
  volume~1, pages 169--172, 1997.

\bibitem{Lotfian_Foroosh_2017}
Sina Lotfian and Hassan Foroosh.
\newblock View-invariant object recognition using homography constraints.
\newblock In {\em Proc. IEEE International Conference on Image Processing
  (ICIP)}, 2017.

\bibitem{Milikan_etal_2017}
Brian Milikan, Aritra Dutta, Qiyu Sun, and Hassan Foroosh.
\newblock Compressed infrared target detection using stochastically trained
  least squares.
\newblock {\em IEEE Transactions on Aerospace and Electronics Systems}, page
  accepted, 2017.

\bibitem{millikan2015initialized}
Brian Millikan, Aritra Dutta, Nazanin Rahnavard, Qiyu Sun, and Hassan Foroosh.
\newblock Initialized iterative reweighted least squares for automatic target
  recognition.
\newblock In {\em Military Communications Conference, MILCOM, IEEE}, pages
  506--510, 2015.

\bibitem{Millikan_etal2015}
Brian~A. Millikan, Aritra Dutta, Nazanin Rahnavard, Qiyu Sun, and Hassan
  Foroosh.
\newblock Initialized iterative reweighted least squares for automatic target
  recognition.
\newblock In {\em Proc. of MILICOM}, 2015.

\bibitem{moore2008learning}
Brendan Moore, Marshall Tappen, and Hassan Foroosh.
\newblock Learning face appearance under different lighting conditions.
\newblock In {\em Proc. IEEE Int. Conf. on Biometrics: Theory, Applications and
  Systems}, pages 1--8, 2008.

\bibitem{Morley_Foroosh2017}
Dustin Morley and Hassan Foroosh.
\newblock Improving ransac-based segmentation through cnn encapsulation.
\newblock In {\em Proc. IEEE Conf. on Computer Vision and Pattern Recognition
  (CVPR)}, 2017.

\bibitem{murray1964wpn}
M.~MURRAY, A.B. DROUGHT, and R.C. KORY.
\newblock {Walking Patterns of Normal Men}.
\newblock {\em The Journal of Bone and Joint Surgery}, 46(2):335, 1964.

\bibitem{phillips2002brc}
P.J. Phillips, S.~Sarkar, I.~Robledo, P.~Grother, and K.~Bowyer.
\newblock {Baseline results for the challenge problem of human id using gait
  analysis}.
\newblock {\em Proc. of the 5th IEEE Int. Conf. on Automatic Face and Gesture
  Recognition}, 2002.

\bibitem{pollick2002estimating}
F.E. Pollick, V.~Lestou, J.~Ryu, and S.B. Cho.
\newblock {Estimating the efficiency of recognizing gender and affect from
  biological motion}.
\newblock {\em Vision Research}, 42(20):2345--2355, 2002.

\bibitem{prest2012weakly}
Alessandro Prest, Cordelia Schmid, and Vittorio Ferrari.
\newblock Weakly supervised learning of interactions between humans and
  objects.
\newblock {\em IEEE Transactions on Pattern Analysis and Machine Intelligence},
  34(3):601--614, 2012.

\bibitem{rabiner1989thm}
LR~Rabiner.
\newblock {A tutorial on hidden Markov models and selected applications
  inspeech recognition}.
\newblock {\em Proceedings of the IEEE}, 77(2):257--286, 1989.

\bibitem{schollhorn2002iiw}
WI~Sch{\"o}llhorn, BM~Nigg, DJ~Stefanyshyn, and W.~Liu.
\newblock {Identification of individual walking patterns using time discrete
  and time continuous data sets}.
\newblock {\em Gait \& Posture}, 15(2):180--186, 2002.

\bibitem{shakhnarovich2001ifa}
G.~Shakhnarovich, L.~Lee, and T.~Darrell.
\newblock {Integrated Face and Gait Recognition from Multiple Views}.
\newblock {\em IEEE COMPUTER SOCIETY CONFERENCE ON COMPUTER VISION AND PATTERN
  RECOGNITION}, 1, 2001.

\bibitem{shekarforoush1996super}
H~Shekarforoush.
\newblock {\em Super-Resolution in Computer Vision}.
\newblock PhD thesis, PhD Thesis, University of Nice, 1996.

\bibitem{shekarforoush1995sub}
H~Shekarforoush, M~Berthod, and J~Zerubia.
\newblock Sub-pixel reconstruction of a variable albedo lambertian surface.
\newblock In {\em Proceedings of the British Machine Vision Conference (BMVC)},
  volume~1, pages 307--316.

\bibitem{shekarforoush1998adaptive}
H~Shekarforoush and R~Chellappa.
\newblock adaptive super-resolution for predator video sequences.

\bibitem{shekarforoush2000multifractal}
H~Shekarforoush and R~Chellappa.
\newblock A multifractal formalism for stabilization and activity detection in
  flir sequences.
\newblock In {\em Proceedings, ARL Federated Laboratory 4th Annual Symposium},
  pages 305--309, 2000.

\bibitem{shekarforoush1998multi}
H~Shekarforoush, R~Chellappa, H~Niemann, H~Seidel, and B~Girod.
\newblock Multi-channel superresolution for images sequences with applications
  to airborne video data.
\newblock {\em Proc. of IEEE Image and Multidimensional Digital Signal
  Processing}, pages 207--210, 1998.

\bibitem{shekarforoush1999conditioning}
Hassan Shekarforoush.
\newblock {\em Conditioning bounds for multi-frame super-resolution
  algorithms}.
\newblock Computer Vision Laboratory, Center for Automation Research,
  University of Maryland, 1999.

\bibitem{Foroosh_2000}
Hassan Shekarforoush.
\newblock Noise suppression by removing singularities.
\newblock {\em IEEE Trans. Signal Processing}, 48(7):2175--2179, 2000.

\bibitem{shekarforoush2000noise}
Hassan Shekarforoush.
\newblock Noise suppression by removing singularities.
\newblock {\em IEEE transactions on signal processing}, 48(7):2175--2179, 2000.

\bibitem{shekarforoush1999super}
Hassan Shekarforoush, Amit Banerjee, and Rama Chellappa.
\newblock Super resolution for fopen sar data.
\newblock In {\em Proc. AeroSense}, pages 123--129. International Society for
  Optics and Photonics, 1999.

\bibitem{Foroosh_etal_1996}
Hassan Shekarforoush, Marc Berthod, Michael Werman, and Josiane Zerubia.
\newblock Subpixel bayesian estimation of albedo and height.
\newblock {\em International Journal of Computer Vision}, 19(3):289--300, 1996.

\bibitem{shekarforoush19953d}
Hassan Shekarforoush, Marc Berthod, and Josiane Zerubia.
\newblock 3d super-resolution using generalized sampling expansion.
\newblock In {\em Proc. International Conf. on Image Processing (ICIP)},
  volume~2, pages 300--303, 1995.

\bibitem{shekarforoush1995subpixel}
Hassan Shekarforoush, Marc Berthod, and Josiane Zerubia.
\newblock {\em Subpixel image registration by estimating the polyphase
  decomposition of the cross power spectrum}.
\newblock PhD thesis, INRIA-Technical Report, 1995.

\bibitem{shekarforoush1996subpixel}
Hassan Shekarforoush, Marc Berthod, and Josiane Zerubia.
\newblock Subpixel image registration by estimating the polyphase decomposition
  of cross power spectrum.
\newblock In {\em Proc. IEEE Conf. Computer Vision and Pattern Recognition
  (CVPR)}, pages 532--537, 1996.

\bibitem{shekarforoush1998blind}
Hassan Shekarforoush and Rama Chellappa.
\newblock Blind estimation of psf for out of focus video data.
\newblock In {\em Image Processing, 1998. ICIP 98. Proceedings. 1998
  International Conference on}, pages 742--745, 1998.

\bibitem{Foroosh_Chellappa_1999}
Hassan Shekarforoush and Rama Chellappa.
\newblock Data-driven multi-channel super-resolution with application to video
  sequences.
\newblock {\em Journal of Optical Society of America-A}, 16(3):481--492, 1999.

\bibitem{shekarforoush2000multi}
Hassan Shekarforoush and Rama Chellappa.
\newblock A multi-fractal formalism for stabilization, object detection and
  tracking in flir sequences.
\newblock In {\em Proc. of International Conference on Image Processing
  (ICIP)}, volume~3, pages 78--81, 2000.

\bibitem{shekarforoush1998denoising}
Hassan Shekarforoush, Josiane Zerubia, and Marc Berthod.
\newblock Denoising by extracting fractional order singularities.
\newblock In {\em Proc. of IEEE International Conf. on Acoustics, Speech and
  Signal Processing (ICASSP)}, volume~5, pages 2889--2892, 1998.

\bibitem{Shen2008}
Y.~Shen and H.~Foroosh.
\newblock View-invariant recognition of body pose from space-time templates.
\newblock In {\em Proc. of CVPR}, 2008.

\bibitem{Shen2009}
Y.~Shen and H.~Foroosh.
\newblock View-invariant action recognition from point triplets.
\newblock {\em IEEE Trans. Pattern Anal. Mach. Intell.}, 31(10):1898--1905,
  2009.

\bibitem{shen2008action}
Yuping Shen, Nazim Ashraf, and Hassan Foroosh.
\newblock Action recognition based on homography constraints.
\newblock In {\em Proc. of IAPR Int. Conf. Pattern Recognition (ICPR)}, pages
  1--4, 2008.

\bibitem{shen2008view}
Yuping Shen and Hassan Foroosh.
\newblock View-invariant action recognition using fundamental ratios.
\newblock In {\em Proc. IEEE Conference on Computer Vision and Pattern
  Recognition (CVPR)}, pages 1--6, 2008.

\bibitem{Shen_Foroosh_FR2008}
Yuping Shen and Hassan Foroosh.
\newblock View invariant action recognition using fundamental ratios.
\newblock In {\em Proc. IEEE Conference on Computer Vision and Pattern
  Recognition (CVPR)}, 2008.

\bibitem{shen2008view-2}
Yuping Shen and Hassan Foroosh.
\newblock View-invariant recognition of body pose from space-time templates.
\newblock In {\em Proc. of IEEE Conf. on Computer Vision and Pattern
  Recognition}, pages 1--6, 2008.

\bibitem{Shen_Foroosh_pose2008}
Yuping Shen and Hassan Foroosh.
\newblock View invariant recognition of body pose from space-time templates.
\newblock In {\em Proc. IEEE Conference on Computer Vision and Pattern
  Recognition (CVPR)}, 2008.

\bibitem{Shen_Foroosh_2009}
Yuping Shen and Hassan Foroosh.
\newblock View-invariant action recognition from point triplets.
\newblock {\em IEEE Transactions on Pattern Analysis and Machine Intelligence
  (PAMI)}, 31(10):1898--1905, 2009.

\bibitem{Shu_etal_2016}
Chen Shu, Luming Liang, Wenzhang Liang, and Hassan Forooshh.
\newblock 3d pose tracking with multitemplate warping and sift correspondences.
\newblock {\em IEEE Trans. on Circuits and Systems for Video Technology},
  26(11):2043--2055, 2016.

\bibitem{stevenage1999vag}
S.V. Stevenage, M.S. Nixon, and K.~Vince.
\newblock {Visual analysis of gait as a cue to identity}.
\newblock {\em Applied Cognitive Psychology}, 13(6):513--526, 1999.

\bibitem{sun2014should}
Chuan Sun and Hassan Foroosh.
\newblock Should we discard sparse or incomplete videos?
\newblock In {\em Proceedings of IEEE International Conference on Image
  Processing (ICIP)}, pages 2502--2506, 2014.

\bibitem{sun2011action}
Chuan Sun, Imran Junejo, and Hassan Foroosh.
\newblock Action recognition using rank-1 approximation of joint
  self-similarity volume.
\newblock In {\em Proc. IEEE International Conference on Computer Vision
  (ICCV)}, pages 1007--1012, 2011.

\bibitem{sun2011motion}
Chuan Sun, Imran Junejo, and Hassan Foroosh.
\newblock Motion retrieval using low-rank subspace decomposition of motion
  volume.
\newblock In {\em Computer Graphics Forum}, volume~30, pages 1953--1962. Wiley,
  2011.

\bibitem{Sun_etal_2012}
Chuan Sun, Imran Junejo, and Hassan Foroosh.
\newblock Motion sequence volume based retrieval for 3d captured data.
\newblock {\em Computer Graphics Forum}, 30(7):1953--1962, 2012.

\bibitem{Sun_etal_2015}
Chuan Sun, Imran Junejo, Marshall Tappen, and Hassan Foroosh.
\newblock Exploring sparseness and self-similarity for action recognition.
\newblock {\em IEEE Transactions on Image Processing}, 24(8):2488--2501, 2015.

\bibitem{sun2014feature}
Chuan Sun, Marshall Tappen, and Hassan Foroosh.
\newblock Feature-independent action spotting without human localization,
  segmentation or frame-wise tracking.
\newblock In {\em Proc. of IEEE Conference on Computer Vision and Pattern
  Recognition (CVPR)}, pages 2689--2696, 2014.

\bibitem{sundaresan2003hmm}
A.~Sundaresan, A.~RoyChowdhury, and R.~Chellappa.
\newblock {A hidden Markov model based framework for recognition of humans from
  gait sequences}.
\newblock {\em Image Processing, 2003. Proceedings. 2003 International
  Conference on}, 2, 2003.

\bibitem{tariq2014scene}
Amara Tariq and Hassan Foroosh.
\newblock Scene-based automatic image annotation.
\newblock In {\em Proc. of IEEE International Conference on Image Processing
  (ICIP)}, pages 3047--3051, 2014.

\bibitem{tariq2015feature}
Amara Tariq and Hassan Foroosh.
\newblock Feature-independent context estimation for automatic image
  annotation.
\newblock In {\em Proceedings of the IEEE Conference on Computer Vision and
  Pattern Recognition (CVPR)}, pages 1958--1965, 2015.

\bibitem{Tariq_etal_2017_2}
Amara Tariq, Asim Karim, and Hassan Foroosh.
\newblock A context-driven extractive framework for generating realistic image
  descriptions.
\newblock {\em IEEE Transactions on Image Processing}, 26(2):619--632, 2002.

\bibitem{Tariq_etal_2017}
Amara Tariq, Asim Karim, and Hassan Foroosh.
\newblock Nelasso: Building named entity relationship networks using sparse
  structured learning.
\newblock {\em IEEE Trans. on on Pattern Analysis and Machine Intelligence},
  2017.

\bibitem{tariq2013exploiting}
Amara Tariq, Asim Karim, Fernando Gomez, and Hassan Foroosh.
\newblock Exploiting topical perceptions over multi-lingual text for hashtag
  suggestion on twitter.
\newblock In {\em The Twenty-Sixth International FLAIRS Conference}, 2013.

\bibitem{tenenbaum9fss}
J.B. Tenenbaum and T.~William.
\newblock {Freeman. Separating style and content}.
\newblock {\em Advances in Neural Information Processing Systems}, 9:662--668.

\bibitem{tolliver2003gse}
D.~Tolliver and R.T. Collins.
\newblock {Gait Shape Estimation for Identification}.
\newblock {\em LECTURE NOTES IN COMPUTER SCIENCE}, pages 734--742, 2003.

\bibitem{troje2002dbm}
N.F. Troje.
\newblock {Decomposing biological motion: A framework for analysis and
  synthesis of human gait patterns}.
\newblock {\em Journal of Vision}, 2(5):371--387, 2002.

\bibitem{unuma1995fpe}
M.~Unuma, K.~Anjyo, and R.~Takeuchi.
\newblock {Fourier principles for emotion-based human figure animation}.
\newblock {\em Proceedings of the 22nd annual conference on Computer graphics
  and interactive techniques}, pages 91--96, 1995.

\bibitem{vasilescu2002hms}
M.~Vasilescu.
\newblock {Human Motion Signatures: Analysis, Synthesis, Recognition}.
\newblock {\em INTERNATIONAL CONFERENCE ON PATTERN RECOGNITION}, 16:456--460,
  2002.

\bibitem{vasilescu2002mai}
M.A.O. Vasilescu and D.~Terzopoulos.
\newblock {Multilinear Analysis of Image Ensembles: TensorFaces}.
\newblock {\em LECTURE NOTES IN COMPUTER SCIENCE}, pages 447--460, 2002.

\bibitem{wang2016factorized}
Min Wang, Baoyuan Liu, and Hassan Foroosh.
\newblock Factorized convolutional neural networks.
\newblock {\em arXiv preprint arXiv:1608.04337}, 2016.

\bibitem{weinland2006fva}
Daniel Weinland, Remi Ronfard, and Edmond Boyer.
\newblock {Free viewpoint action recognition using motion history volumes}.
\newblock {\em CVIU}, 104(2-3):249--257, 2006.

\bibitem{wilson1999phm}
A.D. Wilson and A.F. Bobick.
\newblock {Parametric Hidden Markov Models for Gesture Recognition}.
\newblock {\em IEEE TRANSACTIONS ON PATTERN ANALYSIS AND MACHINE INTELLIGENCE},
  pages 884--900, 1999.

\bibitem{yao2012recognizing}
Bangpeng Yao and Li~Fei-Fei.
\newblock Recognizing human-object interactions in still images by modeling the
  mutual context of objects and human poses.
\newblock {\em IEEE Transactions on Pattern Analysis and Machine Intelligence},
  34(9):1691--1703, 2012.

\bibitem{yao2011human}
Bangpeng Yao, Xiaoye Jiang, Aditya Khosla, Andy~Lai Lin, Leonidas Guibas, and
  Li~Fei-Fei.
\newblock Human action recognition by learning bases of action attributes and
  parts.
\newblock In {\em Computer Vision (ICCV), 2011 s}, pages 1331--1338. IEEE,
  2011.

\end{thebibliography}

\end{document}